\documentclass[a4paper,fleqn]{cas-dc}

\usepackage[numbers]{natbib}
\usepackage{array}

\def\tsc#1{\csdef{#1}{\textsc{\lowercase{#1}}\xspace}}
\tsc{WGM}
\tsc{QE}
\tsc{EP}
\tsc{PMS}
\tsc{BEC}
\tsc{DE}

\DeclareMathOperator*{\argmax}{arg\,max}

\usepackage[ruled,vlined]{algorithm2e}
\hypersetup{
  colorlinks=False,
  citecolor=Violet,
  linkcolor=Red,
  urlcolor=Blue}

\usepackage{tabularx}
\usepackage{multirow}
\usepackage{makecell}
\usepackage{subcaption}
\usepackage{utfsym}
\usepackage{xcolor}
\usepackage{soul}

\usepackage[defaultcolor=red,commandnameprefix=always]{changes}
\usepackage[utf8]{inputenc}
\soulregister{\cite}7 
\soulregister{\citep}7 
\soulregister{\citet}7 
\soulregister{\ref}7 
\soulregister{\pageref}7 

\usepackage[switch]{lineno}

\begin{document}
\emergencystretch 3em 
\let\WriteBookmarks\relax
\def\floatpagepagefraction{1}
\def\textpagefraction{.001}
\shorttitle{A hierarchical spatial-aware algorithm with efficient reinforcement learning for human-robot task planning and allocation in production}
\shortauthors{Jintao Xue et~al.}

\title [mode = title]{A hierarchical spatial-aware algorithm with efficient reinforcement learning for human-robot task planning and allocation in production}



\author[1]{Jintao Xue}[]

\affiliation[1]{organization={Department of Civil Engineering, The University of Hong Kong},
                city={Hong Kong SAR},
                }

\author[1]{Xiao Li}[orcid=0000-0001-9702-4153]\cormark[1]
\author[1]{Nianmin Zhang}[]

\cortext[cor1]{Corresponding author: Department of Civil Engineering, Faculty of Engineering, HW6-07, Haking Wong Building, The University of Hong Kong, Pokfulam, Hong Kong, China}

\cortext[0]{This is the accepted manuscript of an article accepted for publication in
\textit{Robotics and Computer-Integrated Manufacturing (Elsevier)}.
The final published version is available at
\url{https://doi.org/10.1016/j.rcim.2025.103159}. Code under development for open release: \url{https://github.com/jintaoXue/Isaac-Production}.}

\begin{abstract}
In advanced manufacturing systems, humans and robots collaborate to conduct the production process. Effective task planning and allocation (TPA) is crucial for achieving high production efficiency, yet it remains challenging in complex and dynamic manufacturing environments.
The dynamic nature of humans and robots, particularly the need to consider spatial information (e.g., humans' real-time position and the distance they need to move to complete a task), substantially complicates TPA.
To address the above challenges, we decompose production tasks into manageable subtasks. We then implement a real-time hierarchical human-robot TPA algorithm, including a high-level agent for task planning and a low-level agent for task allocation. For the high-level agent, we propose an efficient buffer-based deep Q-learning method (EBQ), which reduces training time and enhances performance in production problems with long-term and sparse reward challenges. 
For the low-level agent, a path planning-based spatially aware method (SAP) is designed to allocate tasks to the appropriate human-robot resources, thereby achieving the corresponding sequential subtasks. We conducted experiments on a complex real-time production process in a 3D simulator. The results demonstrate that our proposed EBQ$\&$SAP method effectively addresses human-robot TPA problems in complex and dynamic production processes.
\end{abstract}



\begin{keywords}
 Task planning and allocation \sep Human-robot collaboration \sep Reinforcement learning

\end{keywords}

\maketitle

\section{Introduction}

Recent technological advances - such as artificial intelligence, robotics, simulation, and digital twins - are laying the foundations for the transformation of manufacturing systems from Industry 4.0 to Industry 5.0 \cite{xu2021industry}. Beyond these developments, a critical factor of Industry 5.0 is the redefinition of the relationship between humans and intelligent manufacturing systems, highlighting the importance of human-centricity \cite{leng2022industry}. Human-robot collaboration (HRC) is a key characteristic of human-centric manufacturing systems, meaning that humans and robots work together in a shared workspace to achieve a common goal, leveraging the unique strengths of both humans and robots. However, the effectiveness of HRC largely depends on whether the task can be planned and allocated properly \cite{banziger2020optimizing}.

Task planning and allocation (TPA) plays a key role in modern manufacturing by optimizing the sequence of tasks and allocating appropriate resources to maximize overall efficiency and profitability \cite{cheng2019task, lee2022task, zhao2025safety}. This is particularly important in the context of the HRC-based manufacturing system, as humans and robots can be considered as production participants in a heterogeneous manner, and TPA is to allocate tasks to these heterogeneous agents according to their characteristics. 
Thus, it is essential to enhance the TPA strategies between humans and robots to facilitate effective HRC, which is crucial for achieving high-performance manufacturing systems.

However, the highly dynamic and complex manufacturing environment presents significant challenges. Firstly, multiple human-robot participants increase the difficulty of human-robot TPA. Secondly, real-time updates of spatial position information are crucial when participants need to move around the workspace while performing production tasks. 
Third, the algorithm must ensure real-time human-robot TPA in interactive changing environmental conditions. These factors make traditional TPA methods less effective \cite{bruno1986rule, johannsmeier2016hierarchical}. Although deep reinforcement learning (DRL) shows promise for real-time sequential decision making, it faces challenges in training efficiency and solution quality \cite{eschmann2021reward}.

This paper aims to address the above challenges using our proposed efficient buffer-based deep Q-learning with a path planning-based spatially aware (EBQ$\&$SAP) method, which enables highly efficient human-robot TPA in a complex production environment. First, for the task description stage \cite{liu2016scheduling}, several predefined production tasks are established based on the configurations and processes of the given production lines. These tasks are then decomposed into manageable subtasks, serving as the finest level of instructions for humans and robots. 
Unlike high-level tasks, subtasks featuring low variability are executed in a clear, predefined sequential order, removing the need for prioritization decisions. They can be easily categorized as either human- or robot-specific, emphasizing task-level planning and allocation.

Second, we present a hierarchical human-robot TPA algorithm EBQ$\&$SAP, that can reduce the complexity and dynamics of decision-making through a two-level agent paradigm. 
The high-level agent employs an RL-based, trainable strategy to dynamically handle task planning, selecting the current task at each time step. The low-level agent employs a path-planning strategy to allocate the task to the nearest human and/or robot for the sequential execution of corresponding subtasks.
Specifically, to facilitate time-efficient decision-making without the need to explicitly model long-term, dynamic, real-time production processes, we use DRL due to its effectiveness in continuous decision-making problems. The sparse reward problem in RL \cite{dewey2014reinforcement} is a common challenge, particularly for long-term tasks such as production processes, and it negatively impacts the performance of DRL algorithms. To overcome this problem, we implement an efficient buffer-based Deep Q-Learning (EBQ) method for the higher-level agent. Based on previous replay buffers work \cite{schaul2015prioritized} for experience reuse and batch training in DRL, we reduce the training time and increase the performance by introducing an efficient buffer mechanism that can modify the reward return of episodes and the amount of data, which is detailed in Sec. \ref{sec:buffer}. 
Furthermore, we incorporate the dueling network architecture \cite{wang2016dueling} and Noisy Nets \cite{NoisyNet} to enhance the performance of our algorithm. For neural network design, we utilize the Attention mechanism to process heterogeneous state information and adopt a Transformer-based architecture to enhance the model’s capacity to solve complex problems \cite{vaswani2017attention}. 

For the lower-level agent, we present a path planning-based spatial-aware method (SAP) that allocates tasks to humans and robots while considering additional spatial factors.
Upon receiving task decisions, the low-level agent uses graph-based path planning to compute distances from each human and robot to key task-related working areas. The nearest human and/or robot is allocated to execute the task’s subtasks, with movement time being a critical factor in human-robot navigation within the environment.

In summary, the main contributions of this paper include the following: 1) A novel hierarchical approach EBQ$\&$SAP, to facilitate real-time TPA while considering the spatial information of humans and robots. 2) An efficient training strategy for the DRL-based high-level agent that addresses the challenges of long-term and sparse reward problems, reduces training time, and enhances performance. 3) The adoption of the Attention mechanism with a Transformer architecture, combined with Noisy Nets and the dueling network, to handle heterogeneous state information and improve the capacity of our neural network model. 

The paper is structured as follows. Section 2 reviews relevant work in TPA, HRC, DRL, and graph-based path planning. Section 3 presents our research method, including the hierarchical algorithm EBQ$\&$SAP, the neural network, and the efficient training strategy. Section 4 details the experimental setup and results to validate the effectiveness of our research method. Section 5 highlights the key novelties of our approach, substantiated by experimental evidence, and discusses the study’s limitations. Finally, Section 6 summarizes the algorithm and outlines future research directions.

\section{Literature review}
In this section, we begin by reviewing relevant studies in human-robot collaboration within manufacturing, covering robot configurations, interaction methods, and application scenarios. Next, we investigate research on the TPA problem, focusing on task description, modeling approaches, and algorithm design, and related work on TPA specifically for HRC in manufacturing. Finally, we review related DRL methods, neural network models, and graph-based path planning methods.
\subsection{Human-robot collaboration in manufacturing} \label{sec:HRC}
Technological progress has led to the increasing application of robots in modern manufacturing systems. As robots take on repetitive and risky tasks in the industrial sector, they free human operators to focus on tasks that require dexterity, flexibility, and critical thinking \cite{mukherjee2022survey}. Due to the indispensability of humans and the complementarity of robots to human capabilities, HRC has become a research hotspot in intelligent manufacturing in recent years \cite{jahanmahin2022human}. 

HRC is defined as humans and robots working together in a shared workspace to achieve a common goal, which is a wide-ranging topic. Some Experts analyze and optimize the layout and ergonomics of collaborative robotic work cells \cite{gualtieri2021emerging, cherubini2016collaborative, tsarouchi2017human}. The configurations of robots show high diversity, including collaborative robots (cobots) \cite{zhang2021manual, bi2021safety}, autonomous mobile robots (AMRs) \cite{gong2022toward, krupas2023human}, humanoid robots \cite{malik2024intelligent}, wearable exoskeletons \cite{huysamen2018assessment, nazari2023applied}, and so on. For instance, Gong et al. \cite{gong2022toward} investigate a human-swarm hybrid system where human workers collaborate with a swarm of autonomous guided vehicles (AGVs) to complete multi-agent pickup and delivery tasks. Krupas et al. \cite{krupas2023human} focus on designing a human-centric conceptual framework to implement collaboration aspects into these applications involving unmanned ground and aerial vehicles.

Robots in HRC can be equipped with various sensors, enabling them to perform complex tasks with greater precision. As a result, research on perception-related has become a prominent area of focus \cite{wang2018deep, yao2024task, hietanen2020ar, neto2019gesture, fan2022vision, wang2024data, zheng2025human}. 
Wang et al. \cite{wang2018deep} investigate deep learning as a data-driven technique for continuous human motion analysis and future HRC needs prediction. Zheng et al. \cite{zheng2025human} propose a real-time dual-hand action segmentation algorithm for adaptive robot assistance and in-process quality checking. Hietanen et al. \cite{hietanen2020ar} propose a depth-sensor-based model for workspace monitoring and an interactive Augmented Reality (AR) user interface for safe HRC. Multimodal fusion of heterogeneous perception information is also studied \cite{liu2022multimodal, li2021toward, wang2025deep, wang2024data}. Some research focuses on brain signal control for HRC, though it remains in the early stages of testing \cite{liu2021brainwave, buerkle2021eeg}.

HRC in manufacturing encompasses various applications, including sequential collaborative tasks in assembly \cite{kim2020estimating, johannsmeier2016hierarchical, liu2022multimodal, zhang2021manual, li2021toward, zhang2022reinforcement, hui2024multi, zheng2023video} and welding \cite{wang2019virtual, lu2023human, wang2019modeling}, handling task \cite{peternel2019selective, deng2017hierarchical} and other collaborative tasks \cite{gong2022toward}. For example, Kim et al. \cite{kim2020estimating} investigate a battery assembly task to assess the risk of human hand intrusion into the robot's safety zone. Lu et al. focus on scheduling in human-robot collaborative welding \cite{lu2023human}. Task planning and allocation are key aspects of enhancing efficient collaboration among different applications and robots. Numerous studies have explored this topic within the context of HRC, which will be discussed later. 

\subsection{Human-robot task planning and allocation in manufacturing}
Task planning and allocation involve determining when, which agents, and how a group of agents should work together to achieve a common goal.
Task description involves defining and outlining a task based on the nature of the production process, available resources, demands, and task-related data \cite{cheng2019task}. A clear and accurate task description can significantly streamline and accelerate the TPA process. Liu et al. \cite{liu2016scheduling} propose a quantitative analysis of task granularity, which guides the coarse-grained decomposition of tasks, enabling the creation of multi-objective models and algorithms. Banziger et al. \cite{banziger2020optimizing} present a method for using standardized work descriptions to automate procedure generation for mobile assistant robots in the human-robot TPA problem. Li et al. \cite{li2023knowledge} study semantic-enriched work packages with different task granularities.

The modeling of the TPA process involves representing the production process using various methods while considering the dynamic nature of the TPA process. The choice of method for TPA process modeling is closely linked to the algorithm design. For instance, optimization-based TPA algorithms typically model the TPA process as an optimization problem using Mixed Integer Programming (MIP) \cite{lee2022task}. Some use graphs to model the TPA process with graph-based searching algorithms \cite{johannsmeier2016hierarchical}.
In contrast, some research incorporates machine learning techniques for TPA, such as reinforcement learning, which is often modeled using Markov Decision Processes (MDP) \cite{zhang2022reinforcement}.

The tasks involved in HRC within manufacturing consist of subtasks. It is essential to find a suitable or even optimal distribution of tasks to humans and robots, considering various factors such as cost and makespan. This is referred to as the TPA problem in HRC, which has been an ongoing research \cite{schmidbauer2023empirical, kiyokawa2023difficulty}.
The paradigms of algorithm design for TPA classified into four main categories: (1) rule-based \cite{bruno1986rule}, (2) search-based \cite{johannsmeier2016hierarchical, merlo2023ergonomic}, (3) optimization-based \cite{faccio2024task, lee2022task, ham2021human, banziger2020optimizing, patel2020decentralized, geng2021particle, fontes2023hybrid, ren2023decision}, (4) reinforcement learning (RL)-based \cite{zhang2022reinforcement, yu2021optimizing, chen2022dynamic, mavrothalassitis2023deep, zhao2023ppo, lee2022digital, theodoropoulos2025automated}. 
Rule-based approaches rely on predefined, expert-designed rules. While these methods are easy to implement for simple TPA problems, they are highly dependent on the knowledge and experience of human experts. Furthermore, they are time-consuming to design and are not effective for handling more complex, dynamic tasks.
Search-based methods model the TPA process in a way that simplifies the decision-making process. One common technique is the use of graph-based models, such as AND/OR graphs, where algorithms like A* search are employed to find efficient solutions \cite{johannsmeier2016hierarchical, merlo2023ergonomic}. Although these algorithms are reliable, their performance can deteriorate significantly in large search spaces, making them computationally expensive and less efficient for real-time applications.
Optimization-based approaches often model the TPA process using mathematical formulations, such as  MIP problems, and solve them with optimization algorithms. For instance, Liau et al. \cite{liau2022genetic} develop a TPA model using the GA algorithm for HRC in the mold assembly. Faccio et al. \cite{faccio2024task} frame safety as a constraint in an MIP model, which aims to both maximize productivity in a collaborative work cell and ensure secure human-robot collaboration. While these methods can provide optimal or nearly optimal solutions, they often treat TPA as NP-hard problems, which can be computationally demanding and unsuitable for real-time TPA scenarios.
Reinforcement Learning (RL) approaches model the sequential nature of the TPA process, where agents learn through trial and error, receiving rewards based on their actions. RL shows great promise in tackling complex and dynamic TPA problems. For example, 
Yu et al. \cite{yu2021optimizing} formulate the HRC assembly process as a chessboard-based Markov game model, using a multi-agent deep Q-learning-based approach to optimize completion time.
Zhang et al. \cite{zhang2022reinforcement} propose a human-robot collaborative RL algorithm to optimize the TPA scheme in assembly processes. 
However, RL methods require substantial training time and extensive interaction experience to achieve optimal performance, which can be a significant challenge.

Despite these advances, research on TPA for HRC is still limited, with most studies focusing on assembly tasks using arm robots and fewer addressing other manufacturing scenarios. Moreover, there is a lack of in-depth exploration into the applications of RL for TPA in HRC, despite its potential for solving real-time TPA challenges in complex, dynamic environments.

\subsection{Deep reinforcement learning and graph-based planning}
Reinforcement Learning (RL) is a machine learning technique where an agent learns to maximize its cumulative reward through interactions with the environment. It is applied in various domains, such as gaming \cite{silver2016mastering}, competition \cite{silver2018general}, and large language models\cite{ouyang2022training}. RL is commonly modeled as an MDP, which is described as a tuple of states, actions, rewards, and transitions. RL paradigms can be categorized in various ways, such as value-based, policy-based, online-offline, on-policy, and off-policy. One core family of RL methods is Q-learning \cite{watkins1992q}, which belongs to the value-based paradigms. Derived from the on-policy Sarsa \cite{sutton2018reinforcement}, Q-learning is an off-policy method, meaning it learns the optimal policy independently of the agent's actions. In Q-learning, the agent’s actions are determined using an action-value function, with actions selected via the $\argmax$ method, as the policy.
With the advancement of deep neural networks, the value functions can now be represented by neural networks. Mnih et al. \cite{mnih2013playing} introduced the first deep-learning model for learning control policies from high-dimensional sensory input. Van et al. \cite{van2016deep} addressed the overestimation issue in DQN with Double DQN (DDQN). Schaul et al. \cite{schaul2015prioritized} proposed prioritized experience replay (PER), which prioritizes significant transitions. Wang et al. \cite{wang2016dueling} introduced the dueling network architecture, separating the action-value functions' network, improving learning efficiency. In contrast to the commonly used $\epsilon$-greedy exploration strategy in RL, Noisy Net adds parametric noise to network weights to improve exploration \cite{NoisyNet}. Hessel et al. \cite{hessel2018rainbow} combined several methods to enhance DQN performance.
Another DRL paradigm is policy-based, which explicitly employs a policy network to directly optimize the agent’s strategy function for maximizing expected cumulative rewards. Compared to value-based methods (e.g., Q-learning), policy gradient methods learn optimal behavior by adjusting policy parameters directly. They can also integrate with value or action-value functions to form an Actor-Critic framework. Representative algorithms include DDPG \cite{lillicrap2015continuous}, PPO \cite{schulman2017proximal}, and SAC \cite{haarnoja2018soft}. Unlike Q-learning, Actor-Critic methods support high-dimensional continuous action spaces.

The sparse reward problem \cite{dewey2014reinforcement} is commonly encountered in long-duration tasks where agents receive few or infrequent rewards, making training challenging and requiring careful reward engineering.
Vaswani et al. proposed the Transformer model and attention mechanism \cite{vaswani2017attention}, which helps models focus on the most relevant parts of input data. 
He et al. introduced ResNet \cite{he2016deep}, using residual connections to enable deeper networks by addressing issues of vanishing gradients.

Dijkstra’s algorithm \cite{dijkstra2022note} finds optimal paths in weighted graphs but is computationally intensive due to exhaustive node exploration. A* \cite{hart1968formal} improves efficiency using heuristics to prioritize promising paths, though its performance relies on heuristic quality. D* \cite{stentz1994optimal}, a dynamic A* variant, efficiently replans in changing environments but is complex to implement. Hybrid A* \cite{montemerlo2008junior} incorporates the vehicle's kinematic constraints, combining grid-based search with continuous state modeling to produce feasible, efficient paths for vehicles or robots in complex environments, though it requires careful tuning for highly dynamic settings.

\subsection{Research gaps}
In summary, the above literature review focuses on HRC, TPA, human-robot TPA in manufacturing, and DRL. However, current research has the following limitations:

(1) Current TPA research faces challenges in managing dynamic collaborative tasks, as the flexibility of production processes and real-time spatial information of humans and robots, affecting task movement distances, substantially complicates TPA.

(2) Despite significant progress in DRL, a common problem in manufacturing TPA tasks is the sparse reward problem caused by the long-duration nature of the production process. In such cases, rewards are typically given only upon completion or failure of a task, resulting in a sparse reward problem.

(3) Current neural network designs for DRL-based TPA face challenges in processing heterogeneous input information of the production process, enhancing training efficiency, and ensuring scalability.


\begin{figure*}[htb] 
\centering	
	\includegraphics[width=0.99 \linewidth, height=0.5\linewidth]{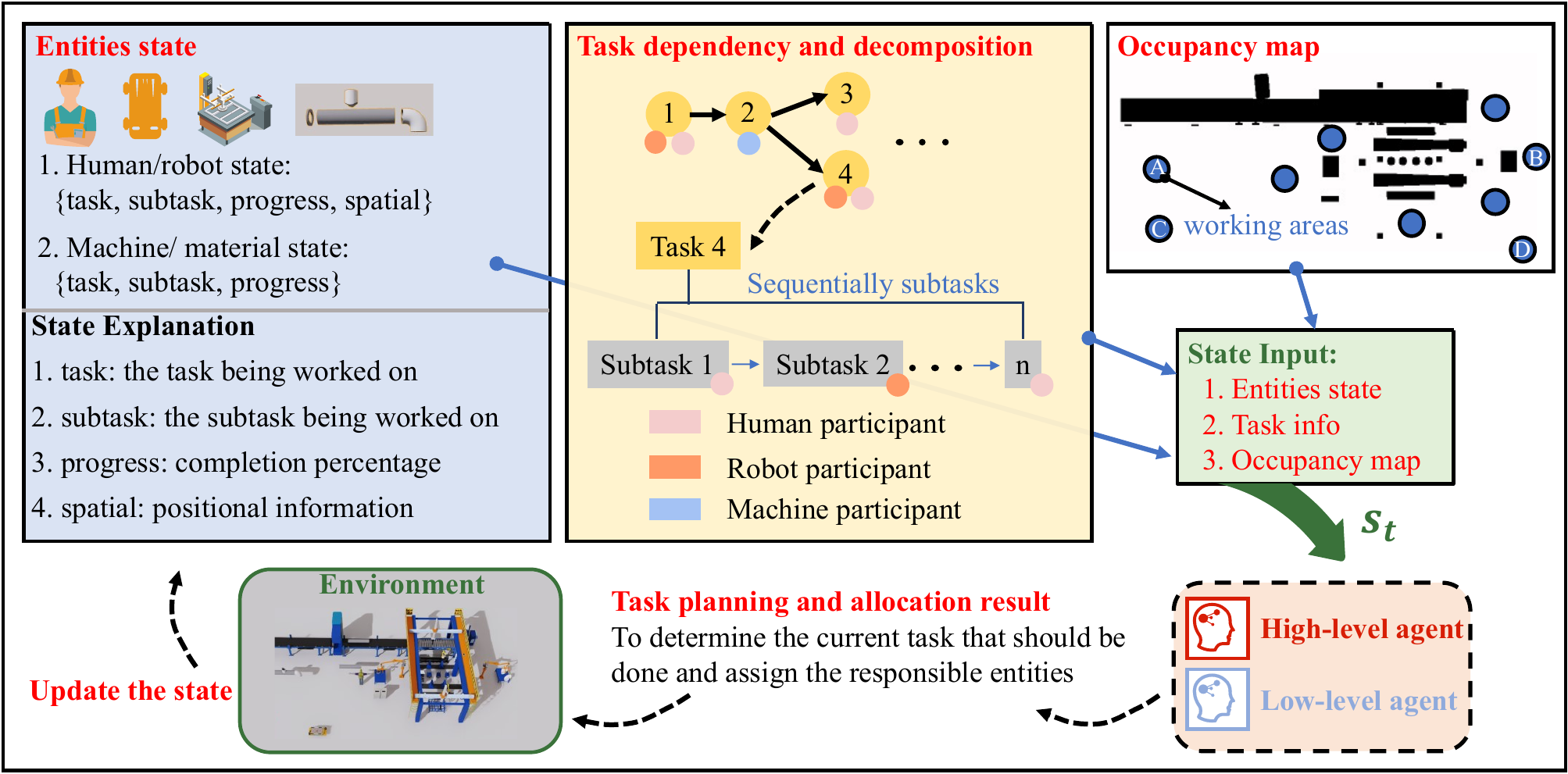}
\caption{Illustration of the real-time production process flow and state information. \label{fig:state_diagram}} 
\end{figure*}

\section{Research method}
To address the challenges mentioned above of TPA in HRC manufacturing, we propose a hierarchical approach EBQ$\&$SAP, that EBQ (Sec. \ref{sec:EBQ}) leverages DRL for task-planning, and SAP (Sec. \ref{sec:SAP}) incorporates spatial data of both humans and robots for task allocation. In addition, an efficient DRL training strategy to improve performance is shown in Sec. \ref{sec:efficient}. In Sec. \ref{sec:architecture}, we adopt the Transformer-based architecture \cite{vaswani2017attention} combined with dueling network \cite{wang2016dueling} and Noisy Net \cite{NoisyNet} to handle heterogeneous information and enhance model capacity.

\subsection{Problem Formulation}\label{sec:problem}
We present a brief description and formulation of the multi-robot-human-machine collaborative production problem. Figure \ref{fig:state_diagram} illustrates the overall production process flow and state information, clarifying the problem formulation. Table \ref{table:notation} explains the notations used.

\subsubsection{Task decomposition}\label{sec:Task}
In the task description phase, a set of predefined production tasks and their dependency graph $\mathcal{T}$ are defined based on the production line’s configurations and processes. Each task is broken down into manageable subtasks, aligned with the task’s nature:
\begin{equation} \label{equation:task}
\begin{gathered}
    \mathcal{T} = \{task_0, task_1, ..., task_i, ...\}, and \,
    \mathcal{T}^{h}, \mathcal{T}^{r}, \mathcal{T}^{m} \subset \mathcal{T} \\
    \mathcal{O} = \{subtask_0, subtask_1, ..., subtask_i, ...\}, \\ 
   and \, \mathcal{O}^{h}, \mathcal{O}^{r}, \mathcal{O}^{m} \subset \mathcal{O},\\
\end{gathered}
\end{equation}
in which $\mathcal{T}$ outlines the high-level tasks for humans, robots, and machines to complete, with subsets $\mathcal{T}^{h}, \mathcal{T}^{r}$, and $\mathcal{T}^{m}$ corresponding to tasks for humans, robots, and machines, respectively. 
$\mathcal{O}$ represents the overall set of subtasks, with
$\mathcal{O}^{h}, \mathcal{O}^{r}$, and $\mathcal{O}^{m}$ provide the most granular level of instruction for humans, robots, and machines to follow step-by-step. 
\begin{equation} \label{equation:task_subtask}
\begin{gathered}
   task_i = \{subtask_0, subtask_1, \ldots, subtask_j, \ldots\},\\
   task_i \in \mathcal{T}, subtask_j \in \mathcal{O},
\end{gathered}
\end{equation}
in which each high-level task $task_i$ has a predefined sequence of subtasks, tailored case-by-case to the production problem’s nature. The subtask batch for each high-level task is simple, requiring only sequential execution. Subtasks are categorized into three classes: $\mathcal{O}^{h}$ (human), $\mathcal{O}^{r}$ (robot), and $\mathcal{O}^{m}$ (machine), indicating which entity should perform the subtask.
\begin{figure*}[htb] 
\centering	
	\includegraphics[width=0.99 \linewidth, height=0.46\linewidth]{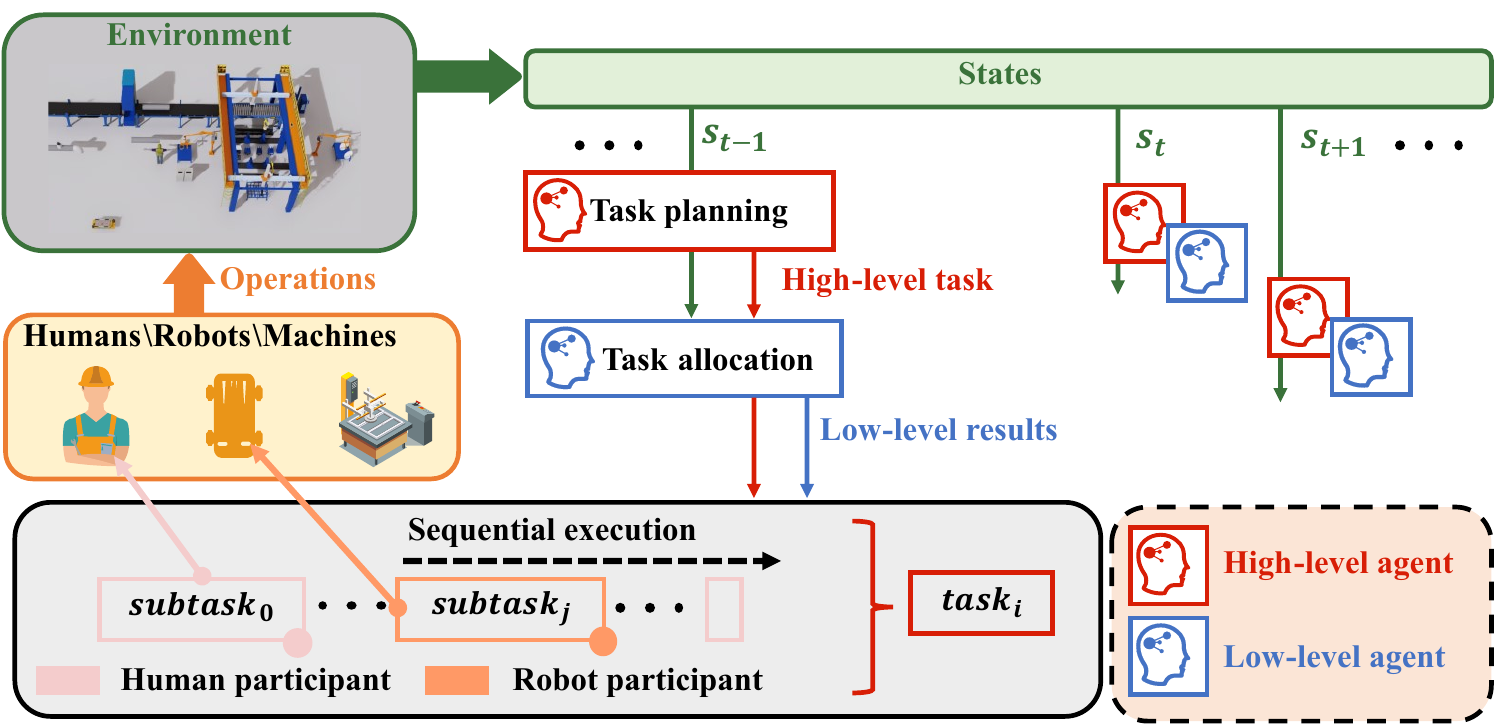}
\caption{Our proposed hierarchical human-robot TPA algorithm with the high-level agent for real-time task planning, and a low-level agent for task allocation. The details of the high-level agent and low-level agent are expressed in Sec. \ref{sec:EBQ} and Sec. \ref{sec:SAP}. \label{fig:framework}} 
\end{figure*}

\subsubsection{Entities state description}\label{sec:entity}
Consider a production line involving groups of operator entities, including robots, humans, and machines, defined as follows:
\begin{equation} \label{equation:entity}
\begin{gathered}
    {\mathcal{E} = \{\mathcal{E}^h, \mathcal{E}^r, \mathcal{E}^{mac}, \mathcal{E}^{mat}\}},
\end{gathered}
\end{equation}
\begin{table}[htb] 
\caption{Notations table.}
\begin{subtable}[ht]{\linewidth}
\centering
\begin{tabular}{p{1cm}<{\centering}p{6.2cm}}
    \toprule 
    \multirow{1}{*}{\makecell[c]{\vspace{0.05cm}Notation}}& \multirow{1}{*}{\makecell[c]{\vspace{0.1cm}Description}} \\ 
    \cline{1-2} \\[-0.5em]  
    {$\mathcal{T}$} & Task set  \\
    {$\mathcal{O}$} & Subtask set   \\
    {$\mathcal{E}$} & Entity set of production participants \\
    {$\mathbf{s}_t$} & State of environment at time step t     \\
    {$\mathbf{R}$} & Reward function   \\
    {$\eta_i$} & Hyperparameters of reward function and buffer, i $\in$ \{1, 2, $\dots$, 6\} \\
    {$\pi$} & Policy of high-level agent     \\
    {$q_{\theta}$} & Q-function parameterized by $\theta$     \\
    {$\theta$} & Parameters of neural network      \\
    \bottomrule 
\end{tabular} 
\end{subtable}
\vspace{-2mm} \label{table:notation}
\end{table}
where $\mathcal{E}$ is the set of all operator entities. The robot set is defined as $\mathcal{E}^r = \{e^r_0, e^r_1, ...,  e^r_i, ...\}$ where $e^r_i$ denotes the $i$-th robot. Similarly, $\mathcal{E}^h = \{e^h_0, e^h_1, ...,  e^h_i, ...\}$ represents human set, and machine set $\mathcal{E}^{mac} = \{e^{mac}_0, e^{mac}_1, ...,  e^{mac}_i, ...\}$. The final product requires various heterogeneous raw materials, represented as \(\mathcal{E}^{mat} = \{e^{mat}_0, e^{mat}_1, ...,  e^{mat}_i, ...\}\), denotes an individual raw material component to be processed. Each product requires these materials to undergo specific processing steps, such as welding or cutting.
The entity states are defined as follows:
\begin{equation} \label{equation:task_subtask}
\begin{gathered}
   \mathbf{s}^h_{i,t}, or \; \mathbf{s}^r_{i,t} = \{task, subtask, progress, spatial\}, \\
    \mathbf{s}^{mac}_{j,t}, or \; \mathbf{s}^{mat}_{j,t}= \{task, subtask, progress\}, \\
    i,j \in \mathcal{E}, task \in \mathcal{T}, subtask \in \mathcal{O}, spatial=\{x,y\},\\
\end{gathered}
\end{equation}
at each time step $t$, the state of a human $\mathbf{s}^h_{i,t}$ or robot $\mathbf{s}^r_{i,t}$ follows a unified format, where $task \in \mathcal{T}$ and $subtask \in \mathcal{O}$ indicate the current task with corresponding subtask allocated to the human or robot. The $progress$ variable reflects the task completion degree based on the task’s nature, while $spatial = \{x, y\}$ provides real-time global position information for the movable human or robot, defined by the global coordinates ${x, y}$. Similar to the human/robot state format, the machine/material state formats exclude spatial information, as it is not critical for this production problem.

\subsubsection{State description and objective function}\label{sec:obj}
The production process is time-dependent, and the state of the production process at time step \( t \) is defined as follows:
\begin{equation} \label{eq:state}
\begin{gathered}
    \mathbf{s}_{t} = \{\mathbf{s}^h_t, \mathbf{s}^r_t, \mathbf{s}^{mac}_t, \mathbf{s}^{mat}_t, \mathcal{T}, \mathcal{O}, Grid\_Map\}, \\
    \mathbf{s}^{[\cdot]}_{t} = \{\mathbf{s}^{[\cdot]}_{0,t}, \mathbf{s}^{[\cdot]}_{1,t}, ..., \mathbf{s}^{[\cdot]}_{i,t}, ...  \}, \\ where [\cdot] \in \{h, r, mac, mat \}, i \in \mathcal{E},
\end{gathered}
\end{equation}
where state \(\mathbf{s}_{t}\) includes the available operator entities state (humans, robots, machines as defined in Sec. \ref{sec:entity}), and task-related information (Sec. \ref{sec:Task}). The \(Grid\_Map\), derived from the raw environment, classifies each grid cell as either free space or occupied (indicating obstacles).
We aim to design an algorithmic function \( F \) to facilitate TPA. This function \( F \) takes state information $\mathbf{s}_t$ as input and outputs a result that directs entities, especially humans and robots, to perform suitable tasks and subtasks at each time step. The TPA function is defined as follows:
\begin{equation} \label{eq:obj}
\begin{gathered}
    allocation_t = F(s_t) =\\
    \{task, e_i^h \; and/or \; e_j^r \}, e_i^h \in \mathcal{E}^h, e_j^r\in \mathcal{E}^r,\\
    s_{t+1} = Env(allocation_t), \; if \; not\; end, \\
    makespan = t, \; if \; end,
\end{gathered}
\end{equation}
where the \( allocation_t \) defines the prioritized task and the responsible operator entities. Subtasks for each task (see Equation \ref{equation:task_subtask}) are predefined, exhibit low variability, and are executed sequentially. The production environment \( Env \) is updated by \( allocation_t \), yielding a sequence of decisions: \( allocation_0, allocation_1, \ldots, allocation_{t} \). The objective is to minimize the total production time, or \( makespan \).

\subsection{Hierarchical task planning and allocation algorithm} \label{sec:framework}
We propose a hierarchical TPA algorithm EBQ$\&$SAP for HRC to address the problem outlined in Section \ref{sec:problem}. As illustrated in Figure \ref{fig:framework}, the algorithm consists of two components: a high-level agent responsible for task planning and a low-level agent focused on task allocation. 
\begin{algorithm*}[htb]

\small \caption{EBQ Algorithm for high-level agent.}\label{alg:efficientDQN}
\SetKwInOut{Input}{input}\SetKwInOut{Output}{output}
\Input{$\mathcal{D}$ -- empty prioritized replay buffer; $\mathcal{D}'$ -- temporary buffer; $\theta$ -- behavior network, $\overline{\theta}$ -- target network}
\Input{$N_r$ -- replay buffer maximum size; $N_b$ -- training batch size; $\overline{N}$ -- target network replacement freq}
\Input{$N_{train}$ -- start training step; $N_{p}$-- buffer replay period; $\eta_{4}, \eta_{5}, \eta_{6}$--hyperparameters in Sec. \ref{sec:reward} and \ref{sec:buffer}}
\Input{$H$ -- time horizon of one episode, $M$ -- episodes maximum size, $t_{total}$ total steps = 0}
\For{$\textit{episode} \in \{ 1, 2, \ldots, M \}$} {
  Initialize temporary buffer $\mathcal{D}'$ = $\emptyset$, time step t = 0  \tcp*{Buffer details in Sec. \ref{sec:buffer} }
  \While{\textit{episode} \textup{is not end}} {
    Observe $s_t$ and choose action $a_t \sim \pi_{\theta}$ \tcp*{State description in Sec. \ref{sec:obj} }
    Sample next state $s_{t+1}$ from environment given $(s_t,a_t)$ and receive reward $r_{t+1}$ \tcp*{MDP in Sec. \ref{sec:MDP} }
    Add transition tuple $(s_t, a_t, s_{t+1}, r_{t+1})$ to $\mathcal{D}'$, set $t \leftarrow t+1$ \;} 
  {\bf if} successfully accomplish task {\bf then:} $r_{modified}$ = $\eta_{4}, n_{repeat}=\eta_{6}$, {\bf else:} $r_{modified}$ = -$\eta_5, n_{repeat}=1$\;
  \For{$t \in \{ 0, 1, \ldots, |\mathcal{D}'|\}$}{    
  \For{$k \in \{ 0, 1, \ldots, n_{repeat}\}$}{ Get transition tuple $(s_t, a_t, s_{t+1}, r_{t+1})$ = $\mathcal{D}'[t]$, set $t_{total}$ = $t_{total}$ + 1 \;
    Add modified transition tuple $(s_t, a_t, s_{t+1}, r_{t+1} + r_{modified})$ to $\mathcal{D}$,
    \\~~~~replacing the oldest tuple if $|\mathcal{D}| \ge N_r$ \;
    {\bf if} $t_{total} < N_{train}$ and $(t_{total} \mod (N_{p}-1)) \not\equiv 0 $ {\bf: continue}; \\
    \For{$i \in \{0, 1, \ldots,  N_{p}-1\}$}{
        Sample a minibatch of \( N_b \) tuples \((s_t, a_t, s_{t+1}, r_{t+1}) \sim \mathcal{D}\), \\
        ~~~~along with their importance-sampling weights \(\omega\), determined by the data-normalized priorities $p$\;
        Update behavior work $\theta$ by doing gradient descent step with loss: \\
        ~~~~$L(\theta) = \mathbb{E}_{(s_t,a_t,s_{t+1},r_{t+1}) \sim \mathcal{D}} \left[\left( \omega \left( r_{t+1} + \gamma \argmax_{a'} q_{\overline{\theta}}(s_{t+1},a')  - q_{\theta}(s_t,a_t) \right) \right)^{2} \right]$ \tcp*{Loss function in Sec. \ref{sec:loss} }
        Replace target network $\overline{\theta} \leftarrow \theta$ every $\overline{N}$ steps\;
    } }

    }
}
\end{algorithm*}
The production environment operates in real-time, with the algorithm receiving state information, denoted as \(s_t\), at each time step. Directly inputting the production goal and outputting the subtask allocation result is complex and difficult to model. Therefore, we first have the high-level agent determine which task, \(task_i \in \mathcal{T}\), should be prioritized based on its potential impact on the overall production makespan. When a task is chosen, the high-level agent transfers the decision to the low-level agent, which allocates the task to suitable operators (human or robot). Subtasks adhere to a clear, predefined sequential order, negating the need for prioritization decisions. They are straightforwardly classified as either human- or robot-specific, with the main emphasis on allocating tasks to the appropriate human or robot within a multi-human, multi-robot environment.

The strengths of our algorithm are evident. First, it is designed for real-time decision-making, with the ability to adapt to dynamic environmental changes. Second, the hierarchical structure simplifies the overall problem by allowing the high-level agent to prioritize tasks that need immediate attention. Third, the algorithm is highly flexible, supporting the deployment of different algorithms for both the high-level and the low-level agent, which we will discuss in more detail in later sections.

\subsection{Efficient Deep Q-learning for high-level agent} \label{sec:EBQ}

The output of the high-level agent consists of discretized actions, making it suitable for algorithms from the Deep Q-Network (DQN) family. Therefore, we adopt Q-learning as the base algorithm. Section \ref{sec:buffer} details our mechanism for enhancing training efficiency and performance. Section \ref{sec:obj} presents our objective function and loss function. Section \ref{sec:reward} describes the design of our reward function. Finally, Algorithm \ref{alg:efficientDQN} provides the pseudocode for the detailed training process.

\subsubsection{Markov decision process} \label{sec:MDP}
We model the high-level agent’s decision-making process using a Markov Decision Process, represented by the tuple \( G = \left< \mathcal{S}, \mathcal{A}, T, \mathcal{R}, \rho_0, \gamma, H \right> \). Here, \( \mathcal{S} \) denotes the state space of the agent, and \( \mathcal{A} \) represents the finite action space. The state space \( \mathcal{S} \) is fully observed by the agent, so we omit the observation space. The reward function is denoted by \( \mathcal{R} \), and \( T \) is the state transition function. \( \rho_0 \) represents the initial state distribution, \( \gamma \in (0, 1] \) is the discount factor, and \( H \) is the time horizon. At each time step \( t \), the agent observes the state \( s_t \in \mathcal{S} \), gets a reward $r_t \in \mathcal{R}$, and selects an action \( a_t \in \mathcal{A} \). The environment then transitions to the next state \( s_{t+1} \) according to the state transition function that describes the probability of transitioning from state \( s_t \) to state \( s_{t+1} \) when action \( a_t \) is taken \( T(s_t, a_t, s_{t+1}) = P[s_{t+1}|s_t, a_t] \), where \( T: \mathcal{S} \times \mathcal{A} \times \mathcal{S} \to [0, 1] \). The agent also receives a reward \( r_{t+1} \in \mathcal{R} \) corresponding to the new state. 

The total accumulated discounted reward from the current time step can be calculated as:
$G_t = \sum_{k=0}^{H} \gamma^k r_{t+k+1}$.
The agent’s objective is to find an optimal action policy \( \pi_\theta \) that maximizes the discounted return. The action-value function (Q-function for short) \( q_\theta(s_t, a_t) \) represents the expected return given a state \( s_t \) and action \( a_t \), and is defined as: 
$q_\theta(s_t, a_t) = \mathbb{E}_\pi [G_t | s_t, a_t, \theta]$.
This function helps the agent evaluate a particular action when taken in a specific state under the policy \( \pi_\theta \), $\theta$ is the policy's network parameter.

\subsubsection{Reward function} \label{sec:reward}
The original reward design is based on achieving the production goal within the shortest makespan, with the constraint that the makespan should not exceed the maximum time horizon 
$H$. Therefore, we propose incorporating three factors into the reward function as follows:
\begin{equation} \label{equation:reward}
\begin{gathered}
    R = r_{time}+ r_{progress} + r_{goal},\\
    r_{time} = -\eta_1,\\
    r_{goal} = 
    \left\{
    \begin{aligned}
    -\eta_2, \quad  & if\; goal\; not\; done\; till\; end, \\
    \eta_2, \quad & if\; goal\; done\; in\; time\; horizon,\\
    0, \quad & else,\\
    \end{aligned}
    \right. \\
    r_{progress} = 
    \left\{
    \begin{aligned}
    \eta_3, \quad & if\, make\, new\, progress,\\
    0, \quad & else.\\
    \end{aligned}
    \right.
\end{gathered}
\end{equation}
where \( \eta_1, \eta_2, \eta_3 \) are hyperparameters, each defined as constant floating-point values. The term \( r_{time} \) provides a negative penalty as time progresses, reflecting the goal of minimizing makespan. If the production goal is completed ahead of schedule, within or before the defined time horizon \( H \), a positive reward \( r_{goal} \) is given. Conversely, a negative reward is applied if the goal is not achieved by the end of the time horizon. For \( r_{progress} \), if progress is made toward the production goal—i.e., products are produced, even if the overall goal is not yet met—a positive reward is provided to encourage continuous advancement toward the final objective.

The most important goal is to complete the production task within the given time horizon. Therefore, \( r_{goal} \) should be the primary reward term. However, \( r_{goal} \) only appears once during the entire production process: either at the end of the time horizon or when the goal is completed, making it a sparse reward.
As a result, the policy may easily learn a suboptimal strategy based on the other two reward terms—\( r_{time} \) and \( r_{progress} \)—with little contribution from \( r_{goal} \) due to its infrequent occurrence compared to the other rewards. We then propose reward shaping with an efficient experience replay buffer to overcome the sparse reward problem. 

\subsubsection{Reward shaping with efficient experience replay buffer} \label{sec:buffer}
During the training phase, transitions obtained through interaction with the environment are represented as a tuple \( (s_t, a_t, s_{t+1}, r_{t+1}) \). These tuples are stored in the experience replay buffer $\mathcal{D}$, which is used for sampling and updating the policy network. The experience replay buffer utilizes prioritized experience replay (PER), which is a commonly adopted mechanism in DRL-based algorithms and applications \cite{schaul2015prioritized, pan2022understanding}. 

In this paper, we modify the PER mechanism to enhance training efficiency and performance, addressing the sparse reward nature of the production problem discussed in Sec. \ref{sec:reward}.
As shown in Fig. \ref{fig:buffer}, the modification, called episode-end processing, changes how transitions are stored. Rather than immediately storing transitions in the buffer, we first place them into a temporary buffer $\mathcal{D}^{'}$ during the episode. 
At the episode's end, we evaluate the \( r_{goal} \) term, allocate additional rewards, duplicate positive transitions, and store all transitions in the temporary buffer \(\mathcal{D}'\). 
These modified transitions are then retrieved from \(\mathcal{D}'\) and stored in \(\mathcal{D}\) as \((s_t, a_t, s_{t+1}, r_{t+1} + r_{modified})\), where \(r_{modified}\) represents the adjusted reward, which modify the retrieval priority of transitions in $\mathcal{D}$. And calculation principle is shown as follows:

\begin{figure}[htb] 
\centering	
	\includegraphics[width=0.99 \linewidth, height=0.72\linewidth]{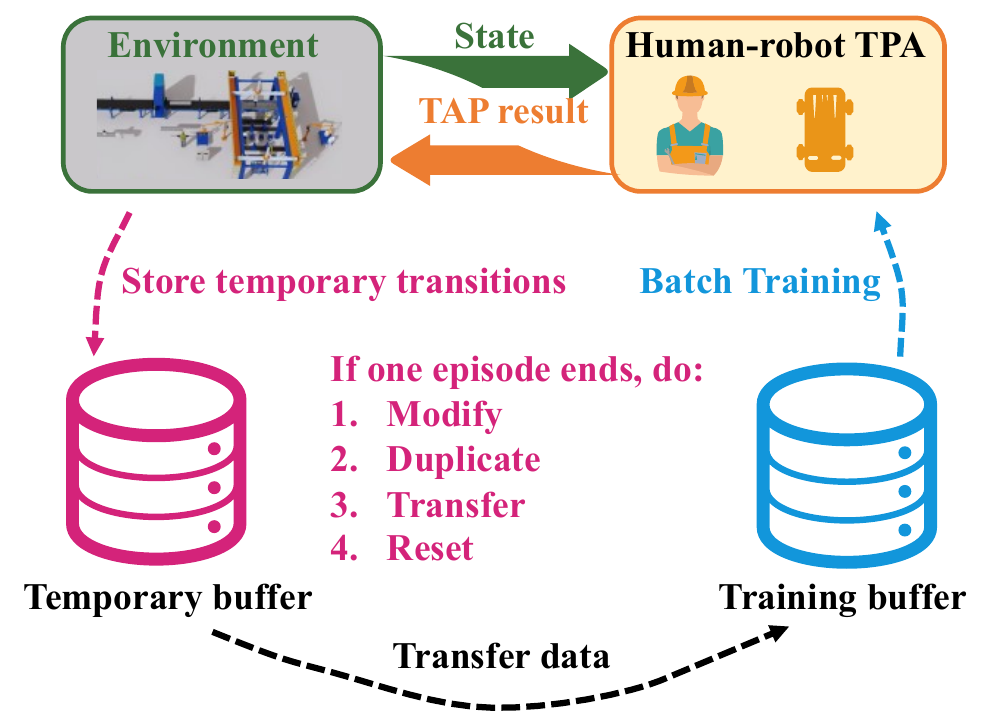}
\caption{Efficient buffer mechanism: episode-end processing. We modify the rewards of each episode's transitions and duplicate the positive data before transferring it to the training buffer, increasing and prioritizing positive data, and shortening training time. \label{fig:buffer}} 
\end{figure}
\begin{equation} \label{eq:modified}
r_{modified} =    
    \left\{
    \begin{aligned}
    \eta_4*(H-makespan)/makespan, \quad & if \; success,\\
    -\eta_5, \quad & if fail,\\
    \end{aligned}
    \right.
\end{equation}
where \( H \) is the maximum time horizon and \( t_{\text{end}} \) is the end time of one episode. If the mission is successful at the end of the episode, we apply a positive modification to all transitions in that episode. Otherwise, a negative penalty is applied. \( \eta_4 \) and \( \eta_5 \) are hyperparameters that remain fixed during training. In practice, we set \( \eta_4 = 0.4 \) and \( \eta_5 = 0.001 \), which work well in our experiments. 
In addition, we duplicate the successful transitions from each episode and store them in the buffer. This mechanism improves data acquisition efficiency, increases the percentage of positive data, and shortens training time. The number of duplicates is \( n_{repeat} = \eta_{6} \), where \( \eta_{6} \) is set to 5 in practice. 
Building on PER, we incorporate the modified reward \( r_{modified} \) from our episode-end processing approach to calculate an adjusted temporal difference (TD) error. Transitions in \(\mathcal{D}\) are then sampled with probabilities proportional to the TD error from their most recent encounter, as detailed below:

\begin{figure*}[htb] 
\centering	
	\includegraphics[width=0.99 \linewidth, height=0.6\linewidth]{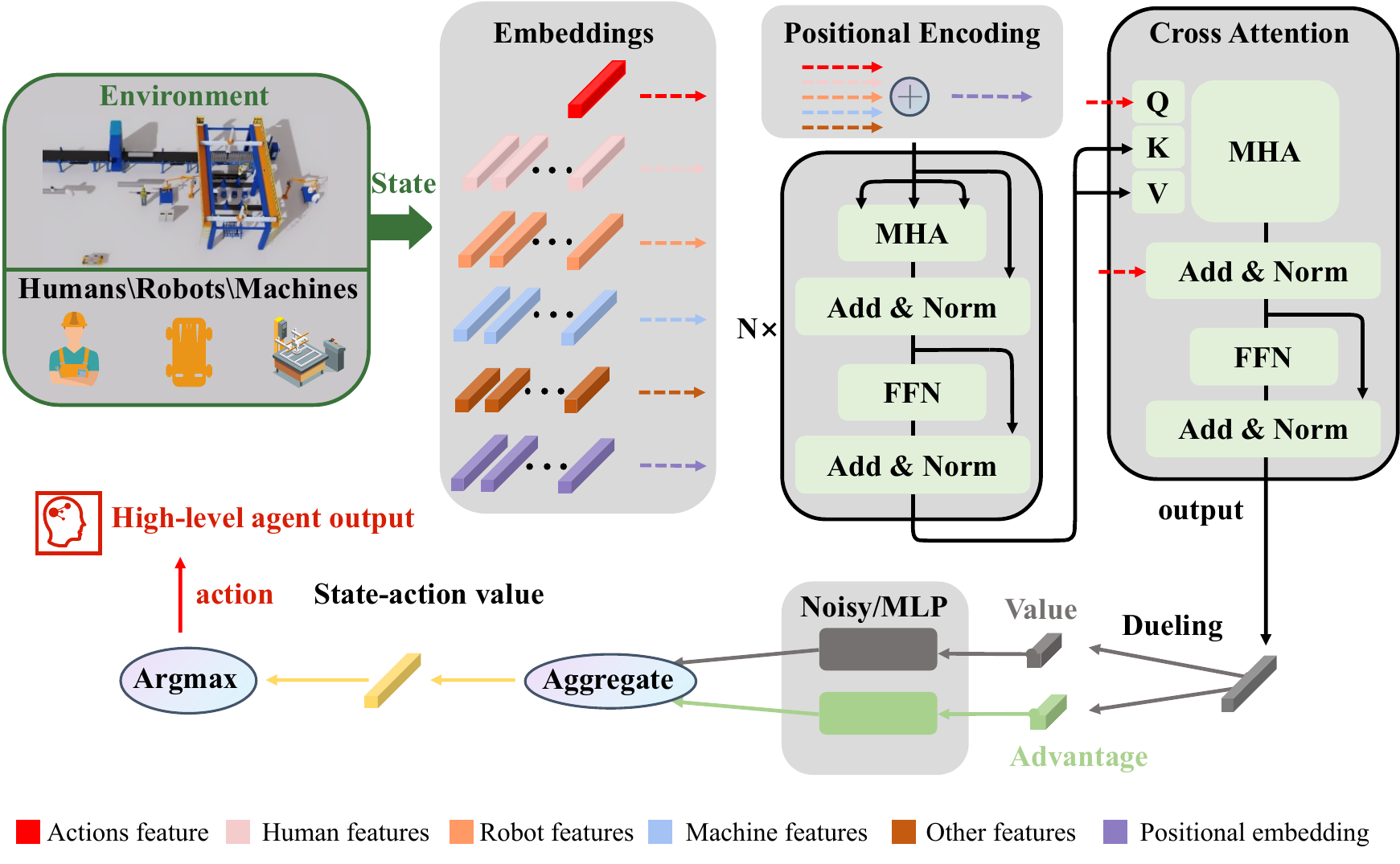}
\caption{Network architecture for high-level agent, including feature embedding, transformer architecture, dueling network and Noisy Net.\label{fig:architecture}} 
\end{figure*}

\begin{equation}
p_t \propto \left| r_{modified} + r_{t+1} + \gamma \max_{a_{t+1}} q_{\overline{\theta}}(s_{t+1}, a_{t+1}) - q_\theta(s_t, a_t) \right|^{\omega} \, ,
\end{equation}
where \( \omega \) is a hyperparameter that adjusts the probability distribution. \(r_{modified} \) is the modified reward. New transitions are inserted into the replay buffer with maximum priority, giving more weight to recent transitions. \( q_{\overline{\theta}}(s_{t+1}, a_{t+1}) \) is the Q-function, and \( \overline{\theta} \) represents the target policy's parameter, $\pi_{ \overline{\theta}}$, which is a stabilized version of the behavior policy \( \pi_{\theta} \) used to stabilize training. 

\subsubsection{Objective function and loss} \label{sec:loss}
The goal of the high-level agent, which employs the RL strategy, is to enhance production efficiency. During the training phase, this process is guided by the reward function defined in Sec. \ref{sec:reward}. Consequently, the objective is to maximize the cumulative discounted reward, expressed by the following objective function:
\begin{equation}\label{equation:object-policy}
\begin{gathered}
    \pi^* = \argmax_{\pi} \mathbb{E}_{\pi} \left[ \sum_{k=0}^{H} \gamma^k r_{t+k+1} \right],  \quad i \in \mathcal{I}, 
\end{gathered}
\end{equation}
where \( \pi \) is the Q-learning-based policy derived from the Q-function \( q_{\theta} \) by selecting actions greedily: \( \pi_{\theta}(s) = \arg\max_{a'} q_{\theta}(s, a') \). Consequently, the optimal policy \( \pi^{*} \) can be expressed as: 
\begin{equation}
    \pi^{*}_{\theta}(s) = \arg\max_{a'} q^{*}_{\theta}(s, a')
\end{equation}
The optimal Q-function \( q^{*}_{\theta} \) is recursively computed using dynamic programming as follows:
\begin{equation}\label{equation:object-q} 
\begin{gathered}
    \; q^{*}_{\theta}(s, a) = \argmax_{\pi_{\theta}} q_{\pi_{\theta}}(s,a). \\ 
    q^{\pi_{\theta}}(s,a)  = \mathbb{E}_{s'} \left[ {r + \gamma \mathbb{E}_{a'\sim\pi_{\theta}(s')}\left[ {q^{\pi_{\theta}}(s',a')}~|~s,a,\pi_{\theta}\right]} \right].
\end{gathered}
\end{equation}
However, since the state-action value function is parameterized by a neural network, we cannot use the recursive equation directly. Instead, we define a loss function for the network and apply gradient descent to update the state-value function:
\begin{equation}\label{eq:loss}
\begin{gathered}
q_{target} = r_{t+1} + \gamma q_{\overline{\theta}}(s_{t+1}, \argmax_{a'} q_{\theta}(s_{t+1}, a')). \\
L(\theta) = \mathbb{E}_{(s_t,a_t,s_{t+1},r_{t+1}) \sim \mathcal{D}} \left[\left( \omega \left( q_{target}  - q_{\theta}(s_t,a_t) \right) \right)^{2} \right],\\
\nabla_{\theta}L(\theta)  = \hfill \\ \mathbb{E}_{(s_t,a_t,s_{t+1},r_{t+1}) \sim \mathcal{D}} \left[ \omega \left( q_{target}  - q_{\theta}(s_t,a_t) \right) \nabla_{\theta} q_{\theta}(s_t,a_t) \right], \\
\end{gathered}
\end{equation}
where \(q_{target}\) is computed using a target Q-function \(q_{\overline{\theta}}\), which is a separate network parameterized by \(\overline{\theta}\). Without this target network, directly updating the network parameters could lead to instability and poor performance due to the recursive nature of Q-learning. To address this, the target network is updated at a lower frequency than the online network $\theta$, which handles interactions with the environment. To minimize the discrepancy between the online and target networks, the loss function is formulated as a mean squared error (MSE), which measures the difference between the Q-values predicted by the online network and the target Q-values. The gradient term, \(\nabla_{\theta}L(\theta)\), is then calculated based on this loss, allowing for an update to the parameters of the online network.

\begin{figure*}[htb] 
\centering	
	\includegraphics[width=0.99 \linewidth, height=0.45\linewidth]{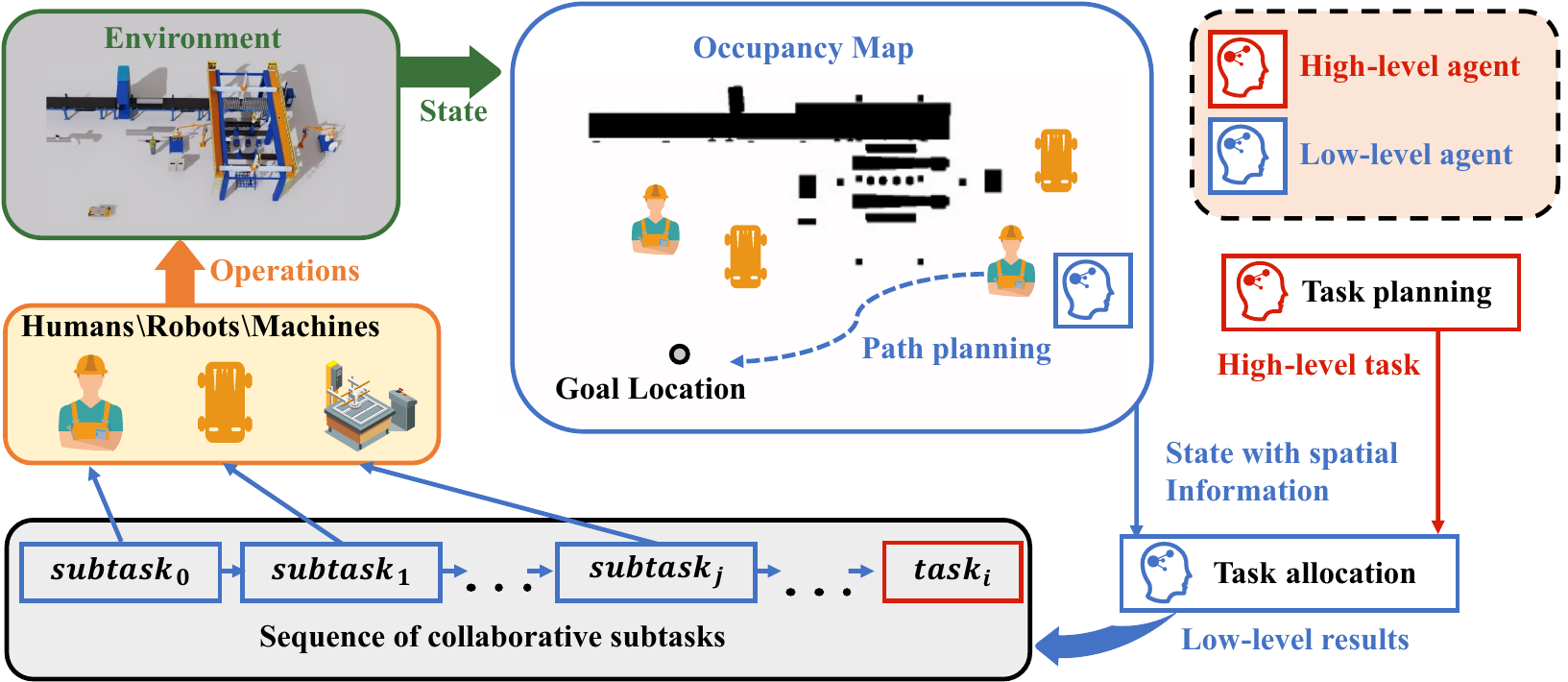}
\caption{SAP: Spatial aware by path planning for low-level agent.\label{fig:low-level}} 
\end{figure*}

\subsubsection{Efficient Deep Q-learning training} \label{sec:efficient}
Algorithm \ref{alg:efficientDQN} presents our training framework, which iteratively updates the Q-function \( q_{\theta} \) (parameterized by \(\theta\)) using gradient descent. Details of objective and loss functions are provided in Sec. \ref{sec:loss}.  
For data acquisition, we employ two buffers: a temporary buffer \( D' \) and a prioritized experience replay buffer \( D \), as detailed in Sec. \ref{sec:buffer}. The temporary buffer \( D' \) is specifically designed to address the issue of sparse rewards. Both buffers store transition tuples \((s_t, a_t, s_{t+1}, r_{t+1})\), with details of the MDP provided in Sec. \ref{sec:MDP}. 
To stabilize training, a target network \(\overline{\theta}\) is introduced, updated less frequently than \(\theta\). Through iterative updates, the Q-function-based policy \(\pi_{\theta}\) progressively converges toward optimality.

\subsection{Neural network architecture of high-level agent} \label{sec:architecture}
The entire network architecture ($\theta$) of the high-level agent is illustrated in Fig. \ref{fig:architecture}, which includes encoders, Transformer-based architecture, dueling network, and Noisy Net.
\subsubsection{Transformer-based architecture} 
 The process begins with environmental information, which encompasses data related to humans, robots, machines, and other relevant factors such as material processing states and overall production progress. The state information is formatted according to Eq. \ref{eq:state} as: $\mathbf{s}_{t} = \{\mathbf{s}^h_t, \mathbf{s}^r_t, \mathbf{s}^{mac}_t, \mathbf{s}^{mat}_t, \mathcal{T}, \mathcal{O}\}$ where $\mathcal{T}$ is the task set, $\mathcal{O}$ is the subtask set, $\mathcal{E}$ is the entity set. Each of these components is represented as a vector. The next step is to pass these components through multiple encoders, as described by the equation:
\begin{eqnarray}
    Encode(c_i) = \rho_i \left( c_i \right), c_i \in \mathbf{s_t}, \rho_i \subset \rho
\end{eqnarray}
Here, for each component \(c_i \in \mathbf{s}_t\), a specific encoder \(\rho_i\) (which is a part of the overall encoders network \(\rho\)) is used to process the information. In practice, the Multilayer Perceptron (MLP) or Embedding models are employed as encoders due to their strong performance and relatively low parameter complexity.

The input features are embedded into a fixed-dimensional space through encoders and categorized into five groups. These heterogeneous features are then passed into the Multi-Head Attention (MHA) layer \cite{vaswani2017attention}, which handles diverse types of information. The MHA layer performs multiple attention operations, where the attention operator is defined as:
\begin{eqnarray}
    \operatorname{Attention}(Q, K, V)=\operatorname{softmax}\left(\frac{Q K^T}{\sqrt{d}}\right) V,
    \label{equation:architecture-attn}
\end{eqnarray}
in this equation, \(\operatorname{Attention}\) calculates the weighted sum of the values (\(V\)) using the dot product of queries (\(Q\)) and keys (\(K\)). We then adopt the Transformer architecture, which includes both self-attention and cross-attention, enhanced by residual connection \cite{he2016deep}. The self-attention uses all features as keys, values, and queries. After several layers of transformation, the output of self-attention is fed into the key and values of the cross-attention model, with action features used as the query to generate action-related outputs. Notably, the action features are derived from the task set \(\mathcal{T}\), which defines the action space of the high-level agent. The network's final output corresponds to the high-level task selection decision, which is strongly influenced by the action features and other types of information. Thus, the action features are used as the query, denoted as \(Q_r = [q_r] \in \mathbb{R}^{1 \times d_k}\), where \(d_k\) represents the dimension of the key vectors. Meanwhile, the attention mechanism treats the other feature types as keys and values to illustrate how they affect the task decision output.
\subsubsection{Dueling network and Noisy Net} 
The output embedding \(\phi\) from cross-attention is then fed into the dueling network architecture \cite{wang2016dueling}, which consists of two separate streams: the value stream and the advantage stream. This separation explicitly distinguishes the representation of state values from the advantages of state-dependent action, enabling the agent to identify the correct action more efficiently during policy evaluation. This is particularly beneficial when redundant or similar actions are introduced to the learning problem:
\begin{eqnarray}
q_\theta(s,a) = v_{\psi_1}(\phi) + a_{\psi_2}(\phi,a) - \frac{\sum_{a'} a_{\psi_2}(\phi, a')}{N_\text{actions}},
\end{eqnarray}
where \(\phi\) is the embedding derived from the MHA, and \(\psi_1\) and \(\psi_2\) represent the value and advantage streams, respectively. $N_\text{actions}$ is the action dimension, to calculate the average advantage. The final output of the network, \(\theta\), is the task decision \(a = \argmax_{a'} q_{\theta}(s,a')\). During the training phase, to further introduce randomness and improve exploration, we incorporate Noisy Nets into the value stream and the advantage stream \cite{NoisyNet}:
\begin{equation} \label{Noisy}
 \textbf{y} = (\textbf{b}  + \textbf{W}  \textbf{x} ) + (\textbf{b} _{noisy} \odot \epsilon^b + (\textbf{W} _{noisy} \odot  \epsilon^w)\textbf{x} ),
\end{equation}
where $\epsilon^b$ and $\epsilon^w$ are random variables, and $\odot$ denotes the element-wise product. The standard linear layer is $\textbf{y} = \textbf{b} + \textbf{W} \textbf{x}$. However, by Noisy Net, we can get more randomness for exploration, which has been shown to outperform the traditional \(\epsilon\)-greedy approach.

\begin{figure*}[htb] 
\centering	
	\includegraphics[width=0.99 \linewidth, height=0.55\linewidth]{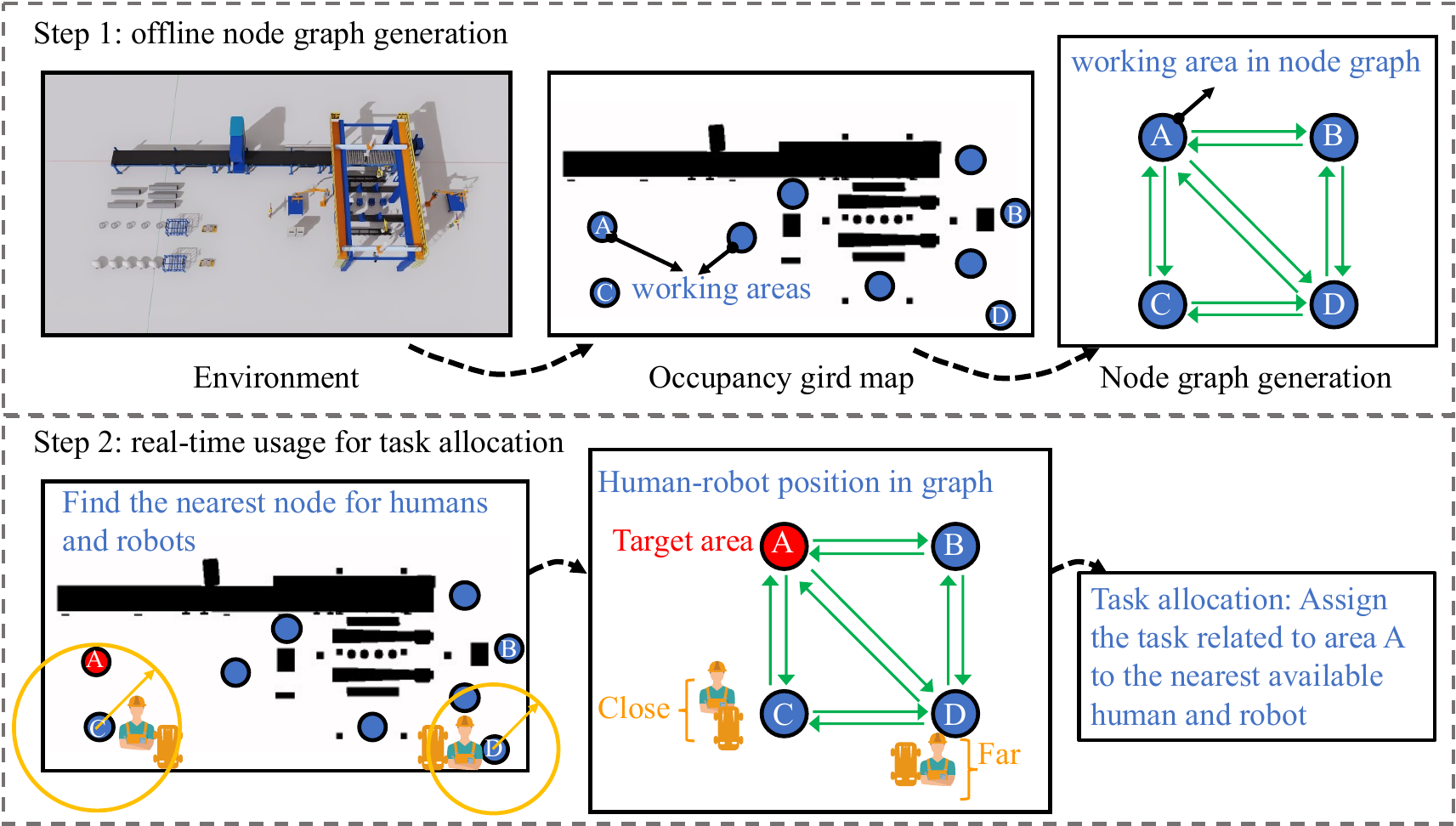}
\caption{SAP: offline node graph generation and real-time usage for task allocation.\label{fig:low_level_method}} 
\end{figure*}
\begin{algorithm*}[htb]

\small \caption{SAP Algorithm for low-level agent.}\label{alg:SAP}

\textbf{\#\#\#Step 1: offline node graph generation} \\
\SetKwInOut{Input}{input}\SetKwInOut{Output}{output}
\Input{$Grid\_Map$ -- occupancy grid map}
Initialize $Node\_Graph$ = $\{\}$ \\
Summarize key working areas as nodes in the $Grid\_Map$, $Area\_nodes = \{node_0, node_1, \ldots \}$, where each node, such as $node_i \in Area\_nodes $, represents a global position with coordinates $node_i = \{x, y\}$. Here, $x$ and $y$ denote the horizontal and vertical coordinates, respectively (Refer to Fig. \ref{fig:low_level_method} for further details); \\

\For{$ node_i \in Area\_nodes$} {
    \For{$ node_j \in Area\_nodes$} {
    {\bf if} $node_j == node_i $ {\bf: continue}; \\
    Do path planning, $Path\_Result = Algo(map = Grid\_Map, start = node_i, goal = node_j)$, where $Algo$ is the path planning algorithm Hybrid A star \cite{gonzalez2015review}; \\
    $Node\_Graph[node_i][node_j]$ = $Path\_Result$
    }
}
\textbf{\#\#\#Step 2: real-time usage for task allocation} \\
\Input{$Node\_Graph$ -- generated node map from step 1; $high\_decision$ -- high-level task decision}
\Input{real time humans and robots global positions}
Initialize $Nodes\_Human$ = $\{\}$, $Nodes\_Robot$ = $\{\}$, -- storing the closes nodes in $Area\_nodes$ for each human and robot.\\

\While{Receive $\; high\_decision$} {
Based on the position information of humans and robots, identify the closest node in $Node\_Graph$ for each, and update $Nodes\_Human$ and $Nodes\_Robot$
accordingly, $Nodes\_Human, Nodes\_Robot \subset Area\_nodes$; \\ Get the most relevant predefined working area node $node_{task} \in Node\_Graph$ based on the high-level task decision; \\
Get the path by $Path = Node\_Graph[node_i][node_{task}]$, where $node_i \in Nodes\_Human$ or $Nodes\_Robot$, and identify the closest human and robot;\\
Allocate the task to the closest human and the robot.}

\end{algorithm*}

\subsection{Spatially-aware low-level agent} \label{sec:SAP}

Figure \ref{fig:low-level} illustrates the collaboration between the low-level and the high-level agent. The low-level agent receives task decisions from the high-level agent and allocates tasks to humans or robots using the SAP algorithm. Tasks consist of predefined subtasks characterized by low variability and a fixed execution sequence, meaning the order of subtask completion does not require real-time decision-making. However, the critical factor is determining which human and/or robot should perform this well-defined sequence of subtasks. This challenge is addressed by our proposed SAP algorithm, which operates in two stages: offline node graph generation and real-time task allocation, as detailed in Algorithm \ref{alg:SAP} and illustrated in Figure \ref{fig:low_level_method}. 
First, an offline node graph, $Node\_Graph$, is generated, precomputing feasible paths and distances using a graph-based path-planning algorithm. Second, during real-time production, the algorithm receives dynamic human and robot global positions, computes task-related distances using the offline node graph, and allocates subtasks to the nearest available human or robot.

\subsubsection{Offline node graph generation} \label{sec:offline}

Movement time is a critical factor in human-robot navigation within an environment. To account for this, we first get the occupancy grid map $Grid\_Map$ from the raw environment, where each grid cell is classified as either free space or occupied (indicating obstacles). Next, we identify key working areas based on the nature of the tasks, as different tasks require distinct working areas. These areas are represented as nodes in a graph, denoted as \( Area\_nodes = \{node_0, node_1, \ldots\} \), where each \( node_i \in Area\_nodes \) is defined by its coordinates \( node_i = \{x, y\} \). $Node\_Graph$ is constructed by finding feasible paths and distances between nodes are precomputed using a graph-based path-planning algorithm, such as Hybrid A* \cite{gonzalez2015review}:

\begin{equation}\label{node_graph} 
\begin{gathered}
    path = Hybrid\_A^*(Grid\_Map, node_i, node_j) =\\
    \{\{x_0, y_0\}, \{x_1, y_1\}, \ldots\},\\
    Node\_Graph[node_i][node_j] = path, \\ 
   Node\_Graph = \Big\{ node_0: \{node_1: path_{0,1}, node_2: \\ path_{0,2}, \ldots\}, node_1:\{node_0: path_{1,0}, \ldots\}, \ldots \Big\},\\
           node_i, node_j \in Area\_nodes,  \\ 
\end{gathered}
\end{equation}
The $Hybrid\_A^*$ path planning algorithm is used to find a feasible path. Given an occupancy grid map $Grid\_Map$, a start node $node_i$, and a goal node $node_j$, the algorithm generates a path as a sequence of coordinates, which is stored in $Node\_Graph$. This results in a comprehensive node graph containing path and distance information between any pair of working areas.

\subsubsection{Real-time task allocation using node graph} \label{sec:online_usage}

Finding paths and calculating distances for humans/robots across all nodes in $Area\_nodes$ using the path-planning algorithm is computationally expensive in real-time task allocation. To address this, we preconstruct a graph of task working area nodes in Sec \ref{sec:offline}, where the available paths and distances between nodes are precomputed using the graph-based path-planning algorithm. With this node graph, the path-planning process is streamlined. 

As detailed in Algorithm \ref{alg:SAP} and illustrated in Figure \ref{fig:low_level_method}, we compute the coordinate distance to identify the nearest node to humans and robots in free space, allocating it as their position, getting $Nodes\_Human, Nodes\_Robot \subset Area\_Nodes$. Distances to other task nodes are then retrieved from the precomputed $Node\_Graph$, eliminating repetitive path planning and significantly reducing computation time. This enables the low-level agent to evaluate spatial distances for potential participants in high-level tasks efficiently.
In addition to spatial information, other factors, such as whether a human is currently available or working, are also considered in the allocation process. The low-level agent selects suitable humans and/or robots for a task, defined as $Task = \{subtask_0, subtask_1, \ldots\}$, with the subtask sequence and responsible entity predefined based on the task's nature. As tasks are sequential and minimally dynamic, no subtask decision planning is required. 

In summary, our SAP algorithm integrates spatial factors to enable efficient human-robot collaboration. Each entity performs its designated role, executing subtasks sequentially to drive production progress toward the overall goal.

\section{Experiments}
In this section, we present the experimental data, including details of the production problem and task decomposition. Following this, we address several key questions. Specifically, we investigate whether the proposed hierarchical algorithm can effectively solve problems in a complex and dynamic production environment and whether the EBQ-based high-level agent is capable of handling task planning. Additionally, we investigate whether the SAP-based low-level agent can manage human-robot task allocation while considering spatial information. We also examine whether our efficient training method, detailed in Sec. \ref{sec:buffer} and Algorithm \ref{alg:efficientDQN}, and the network architecture, described in Sec. \ref{sec:architecture}, can enhance the performance of EBQ.
\begin{figure*}[tpb] 
\centering	
	\includegraphics[width=0.99 \linewidth, height=0.58\linewidth]{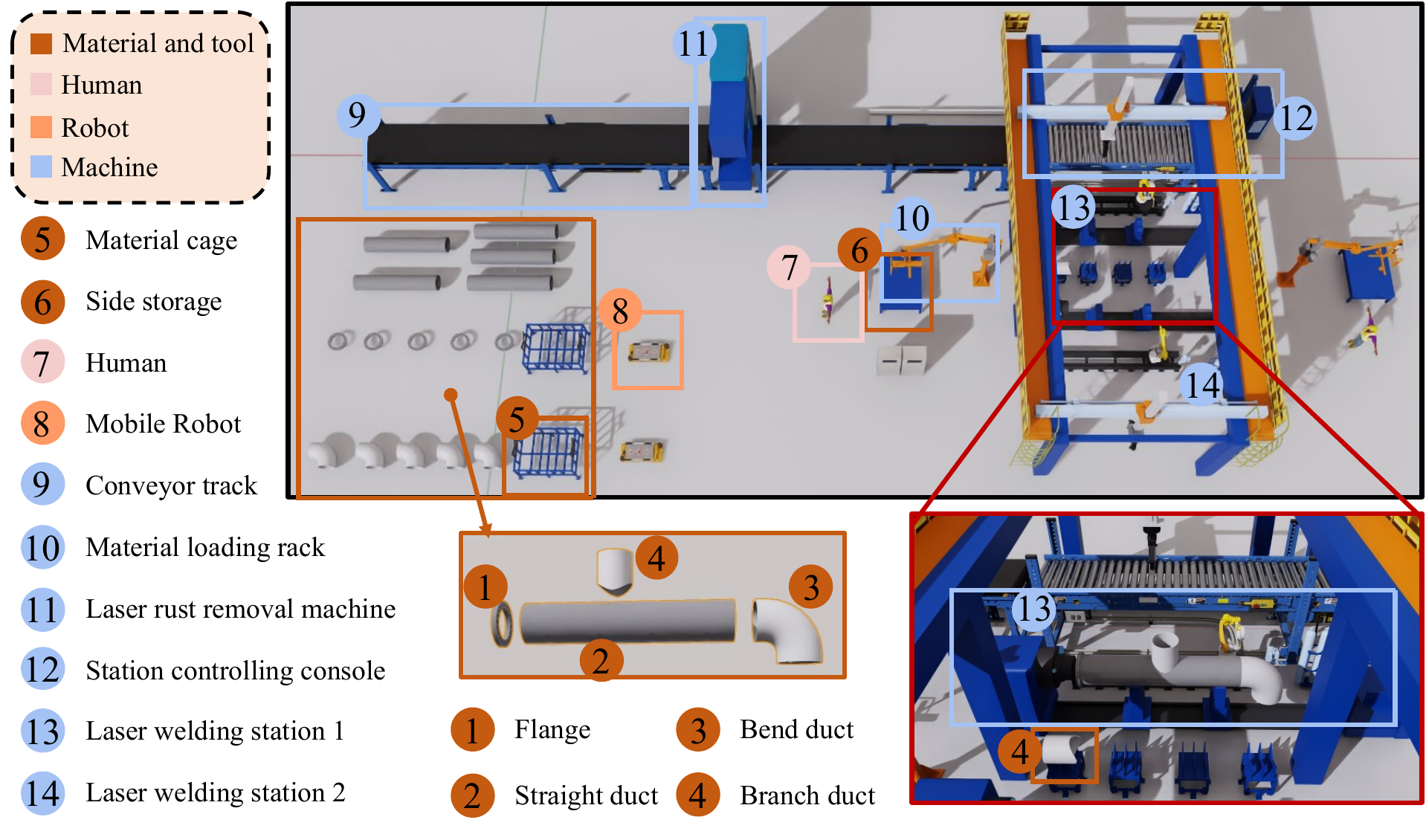}
\caption{Production workspace including humans, robots, machines, and materials.\label{fig:environment}} 
\end{figure*}

\begin{figure*}[htb] 
\centering	
	\includegraphics[width=0.99 \linewidth, height=0.96\linewidth]{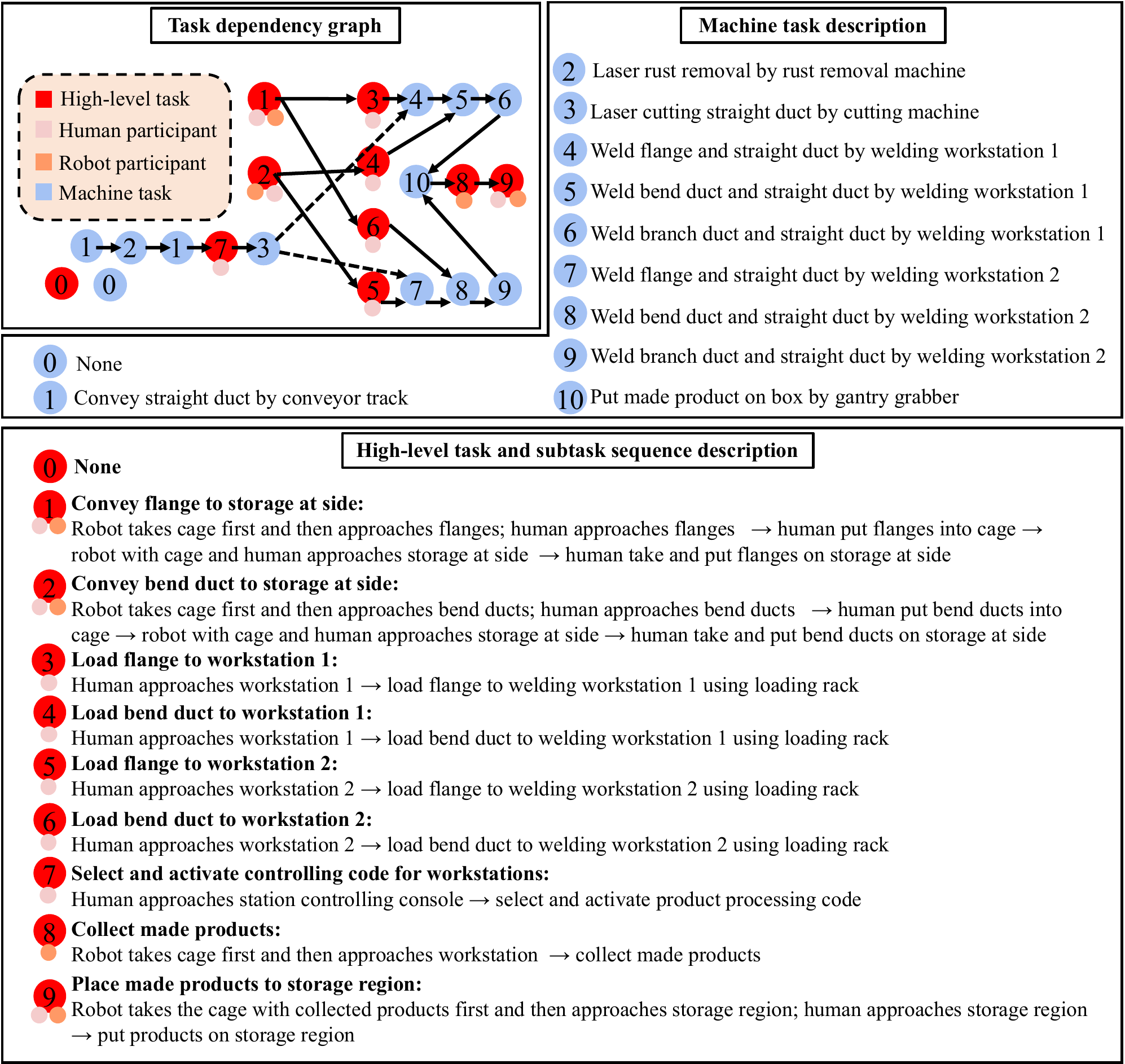}
\caption{Details of the task dependency graph, machine tasks, and human-robot tasks including subtasks. \label{fig:dependency_and_task}} 
\end{figure*}
\subsection{Experiment environment background}

To ensure the realism and relevance of our experiment, we are exploring the manufacturing problem in the context of Modular Integrated Construction (MiC). This modern construction method reduces on-site cast-in-place work by assembling prefabricated components produced in a factory, improving quality, productivity, safety, and sustainability, and reducing labor costs \cite{wuni2020barriers}. The phased production of MiC is essential, and the heavy modular components make it challenging to achieve high efficiency in large-scale production. Therefore, this paper aims to address the challenges in the field of MiC, with a particular focus on improving production efficiency. 

The experimental model is based on a flexible production line for air conditioning ducts in a real MiC factory, with the layout shown in Fig. \ref{fig:environment}. Air conditioning ducts are divided into four components: flanges, straight ducts, bend ducts, and branch ducts. All raw materials are located in the bottom-left region of the environment, except for branch ducts, which are pre-prepared. Due to their weight and the physical effort required to handle them, all materials are challenging to transport manually and require the assistance of mobile robots equipped with material cages to move them to the side storage area. 

The challenge lies in human-robot TPA, which requires the coordination of multiple human operators and mobile robots to ensure a smooth production process.
The simulation is conducted in a 3D simulator, Isaac Sim \cite{liang2018gpu} from NVIDIA.

\subsection{Task dependency graph and decomposition details}
In Sec. \ref{sec:problem}, we formulate the problem. This section provides example data derived from the environment depicted in Fig. \ref{fig:dependency_and_task}, detailing the task dependency graph, machine tasks, and human-robot tasks with their subtasks, highlighting the collaborative roles of humans, robots, and machines in product manufacturing.

\subsection{Comparisons, training protocols, and metrics}
We summarize our comparisons of the efficient buffer mechanism, high-level agent, and network design, and low-level algorithms, with an introduction to training protocols and evaluation metrics.
\subsubsection{Comparisons for Efficient buffer mechanism}

For the efficient buffer-based training method described in Sec. \ref{sec:buffer}, we compare three high-level algorithms: D3QN (Dueling Double DQN \cite{wang2016dueling}), NoModify, and Efficient Buffer Q-learning with Greedy Exploration Strategy (EBQ-G), as shown in Table \ref{table:buffer_comp}. D3QN serves as the baseline, lacking an efficient buffer mechanism, only utilizes prioritized experience replay (PER) \cite{schaul2015prioritized, pan2022understanding}. NoModify partially adopts our efficient buffer approach, incorporating data duplication without the modified reward \( r_{modified} \) obtained at the end of the episode. EBQ-G fully utilizes the efficient buffer mechanism, termed episode-end processing, which modifies transition storage by including the episode reward \( r_{modified} \) and positive data duplication. For comparison, all high-level algorithms utilize the same low-level algorithm: SAP.

\begin{table}[ht]
\caption{Comparison for efficient buffer mechanism proposed in Sec. \ref{sec:buffer}. Our proposed method is indicated by a ``$^\dagger$". \label{table:buffer_comp}}
\centering
\begin{tabular}{p{1.4cm}|c|c} 
\hline
         & \begin{tabular}[p]{@{}l@{}}Modified reward\end{tabular} & \begin{tabular}[c]{@{}l@{}}Positive Data \\ Duplication\end{tabular} \\
\hline
{D3QN} & \scalebox{0.75}{\usym{2613}} & \scalebox{0.75}{\usym{2613}} \\
{NoModify} & \scalebox{0.75}{\usym{2613}} & \checkmark  \\
{EBQ-G}$^\dagger$ & \checkmark & \checkmark \\

\hline
\end{tabular} 
\end{table}

\subsubsection{Comparisons for high-level algorithm and network design}
\begin{table}[ht]
\caption{Comparison for high-level algorithms. Our proposed method is indicated by a ``$^\dagger$"\label{table:charac}. }
\centering
\begin{tabular}{p{1.4cm}|c|c|c|c|c} 
\hline
         & \begin{tabular}[p]{@{}l@{}}Efficient\\buffer\end{tabular} & \begin{tabular}[c]{@{}l@{}} Greedy\end{tabular} & \begin{tabular}[c]{@{}l@{}}Nosiy\\Net\end{tabular} & \begin{tabular}[c]{@{}l@{}}Duel\\Net\end{tabular} & \begin{tabular}[c]{@{}l@{}}Dou-\\ble\end{tabular}\\
\hline
{D3QN} & \scalebox{0.75}{\usym{2613}} & \checkmark & \scalebox{0.75}{\usym{2613}} & \checkmark & \checkmark\\
{EDQN1} & \checkmark & \checkmark & \scalebox{0.75}{\usym{2613}} & \checkmark & \scalebox{0.75}{\usym{2613}} \\
{EDQN2} & \checkmark & \checkmark & \scalebox{0.75}{\usym{2613}} & \scalebox{0.75}{\usym{2613}} & \checkmark \\
{EBQ-G}$^\dagger$ & \checkmark & \checkmark & \scalebox{0.75}{\usym{2613}} & \checkmark & \checkmark \\
{EBQ-N}$^\dagger$ & \checkmark & \scalebox{0.75}{\usym{2613}} & \checkmark & \checkmark & \checkmark \\
{EBQ-GN}$^\dagger$ & \checkmark & \checkmark & \checkmark & \checkmark & \checkmark \\
{EPPO} & \checkmark & \scalebox{0.75}{\usym{2613}} & \checkmark & \checkmark & \checkmark \\
\hline
\end{tabular} 
\end{table}

As shown in Table \ref{table:charac}, for the high-level agent (detailed in Sec. \ref{sec:EBQ}), we evaluate seven algorithms: D3QN \cite{wang2016dueling}, EDQN1, EDQN2 \cite{van2016deep}, EBQ-G, EBQ-N, EBQ-GN, and EPPO \cite{schulman2017proximal}. D3QN enhances Deep Q-learning (DQN) by incorporating Double Q-learning (DDQN) and the dueling network architecture. EDQN1 employs the dueling network but omits Double Q-learning, while EDQN2 uses Double Q-learning without the dueling network. Both EDQN1 and EDQN2 adopt our proposed efficient buffer mechanism. Building on D3QN, EBQ-G integrates the efficient buffer mechanism. EBQ-N replaces the \(\epsilon\)-greedy exploration strategy in EBQ-G with Noisy Nets \cite{NoisyNet}. EBQ-GN combines both \(\epsilon\)-greedy and Noisy Nets exploration strategies. EPPO is a representative Actor-Critic-based algorithm, extending PPO by incorporating our efficient buffer mechanism. For comparison, all the high-level algorithms use the same low-level algorithm: SAP.
\subsubsection{Comparisons for low-level algorithm}

\begin{table}[htb]
\caption{Comparisons for low-level algorithms. See Sec. \ref{sec:SAP} for SAP algorithm details.\label{table:comp_low_level}}
\centering
\begin{tabular}{c|c} 
\hline
         & \begin{tabular}[p]{@{}l@{}}Explanation\end{tabular} \\
\hline
{NoSpatial} & Randomly allocates tasks to entities \\
{LongestPath} & allocates tasks to the farthest entities  \\
{EBQ-G}$^\dagger$ &  allocates tasks to the closest entities\\
\hline
\end{tabular} 
\end{table} 
As shown in Table \ref{table:comp_low_level}, for the low-level agent design detailed in Section \ref{sec:SAP}, we compare three algorithms: NoSpatial, LongestPath, and SAP. NoSpatial lacks path planning-based spatial awareness; upon receiving a high-level task decision, it randomly selects available human and/or robot entities to execute the task and its subtasks. LongestPath incorporates spatial awareness but always chooses the entities farthest from the task location. SAP employs our path planning-based method, updating task-related distances for humans and robots and allocating tasks to the closest human and/or robot. For comparison, all low-level algorithms utilize the same high-level algorithm: EBQ-N.

\subsubsection{Training protocols and metrics}
To simulate real dynamic manufacturing scenarios, the setup includes 1 to 3 workers with the same capabilities and 1 to 3 identical mobile robots, and we randomly initialize the positions of humans and robots in the simulation environment. The human-robot team needs to produce a fixed number of products, which is set to six, in a limited period. During testing, we change the production orders with different amounts of products to test zero-shot performance.
The training and testing stages are set in different random seeds. We utilize the Adam optimizer \cite{kingma2015adam} to optimize all algorithms. All experiments are conducted on a computer with an Intel(R) Xeon(R) Platinum 8370C CPU and NVIDIA GeForce RTX 4090.


For evaluation metrics, we primarily use makespan, progress, and success rate to assess algorithm performance. Progress represents the degree of task completion, where 0 indicates no completion and 1 denotes full completion of the manufacturing order. For efficient buffer mechanism comparison, we include training time as an additional metric. For low-level algorithms, we incorporate distance, which measures the total movement of humans and robots during the production process. Additionally, Gantt charts are used in case studies to evaluate the adaptability and effectiveness of various TPA methods.

\subsection{Performance for efficient buffer mechanism} \label{res:buffer}
\begin{figure*}[htb] 
\centering	
	\includegraphics[width=0.95 \linewidth, height=0.28\linewidth]{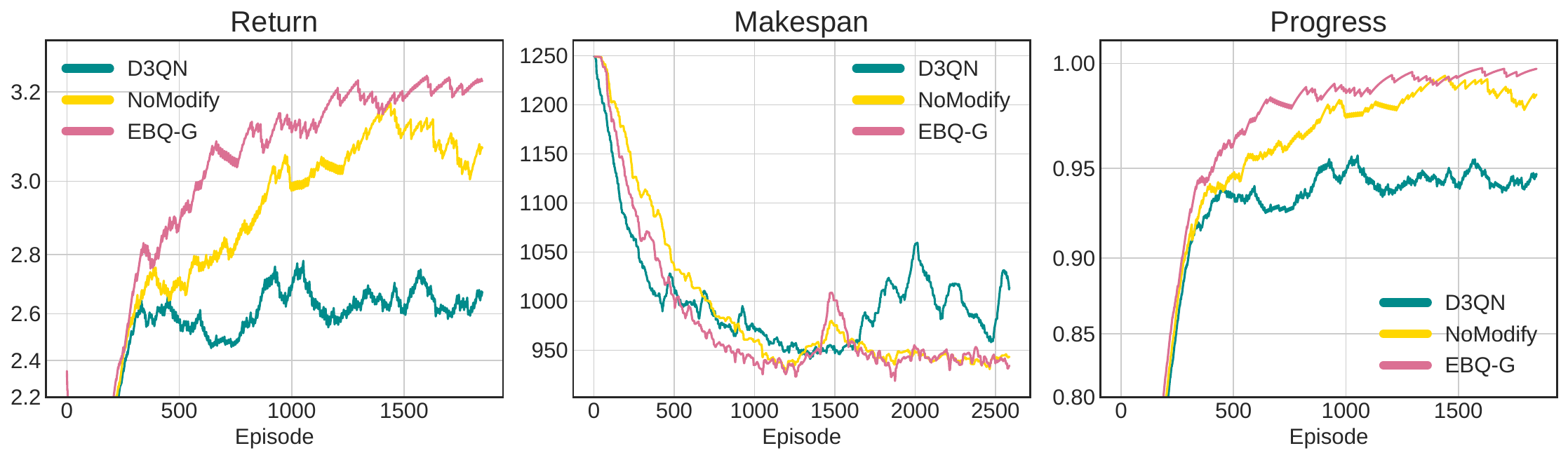}
\caption{Evaluation curves during training intervals for efficient buffer mechanism.\label{fig:buffer_metric}} 
\vspace{-1mm}
\end{figure*}

Firstly, Table \ref{table:training_time} shows the training time required for algorithms. Notably, D3QN does not utilize the positive data duplication, resulting in nearly four times the training time compared to algorithms NoModify and EBQ-G that use the efficient buffer mechanism. 
Figure \ref{fig:buffer_metric} presents the evaluation results at training intervals. D3QN exhibited the worst performance across three metrics: episode return, makespan, and progress. NoModify showed notable improvements in these metrics by adopting positive data duplication compared to D3QN. Table \ref{table:test} illustrates the test stage results for the makespan metric, revealing that EBQ-G achieved a significantly shorter makespan compared to D3QN and NoModify. EBQ-G, which incorporates the episodic reward $ r_{modified} $ in addition to NoModify’s approach, demonstrated further improvements in episode return and progress.
In summary, our proposed episode-end processing method, which allocates additional rewards and duplicates positive data, significantly enhances training efficiency, reduces makespan, and improves task completion degree.
\begin{table}[htb]
\caption{Training time comparison. See Sec. \ref{sec:buffer} for efficient buffer mechanism\label{table:training_time}.}
\centering
\begin{tabular}{c|c} 
\hline
         & \begin{tabular}[p]{@{}l@{}}Training time\end{tabular} \\
\hline
{D3QN} & $\geq$ 40 hours \\
{NoModify} & 10$\sim$11 hours  \\
{EBQ-G}$^\dagger$ & 10$\sim$11 hours \\
\hline
\end{tabular} 
\end{table} 

\begin{table*}[ht] \label{table:test}
\caption{Makespan performance in the testing phase. Note that a "$^\dagger$" indicates the performance of our proposed algorithm. "Hn" denotes the number of humans is set to n, while "Rn" indicates n robots during testing. }

\begin{subtable}[h]{\linewidth}
\centering
\begin{tabular}{b{1.4cm}<{\centering}m{3em}m{3em}m{3em}m{3em}m{3em}m{3em}m{3em}m{3em}m{3em}m{3em}}
\toprule

\multirow{2}{*}[-0.5em]{\makecell[c]{\vspace{0.1cm}Algorithm \\}} & \multicolumn{10}{c}{\normalsize \texttt{Metric: Makespan}}\\

\cline{2-11} \\[-2mm]
& \text{H1,R1} & \text{H1,R2} & \text{H1,R3} & \text{H2,R1}& \text{H2,R2}& \text{H2,R3}& \text{H3, R1}& \text{H3,R2}& \text{H3,R3}& \text{Mean}\\

\cline{1-11} \\[-0.5em]  
{D3QN} & 1013.93 & 983.03  & 961.05& 837.25& 855.96& 746.61&871.47&865.49&737.82 & 874.73 \\
{NoModify} &990.35& 1018.21& 1004.94& 758.72& 789.63& 735.42&825.92 &782.96&702.07&845.36\\
{EDQN1} &1032.35& 1069.95& 1019.08& \textbf{755.75}& 751.14& 752.90& 750.01&692.63&690.46&834.92  \\

{EDQN2} &991.35& \textbf{927.71}&\textbf{922.94}& 758.83& 745.86& 760.86& 755.80&702.56&706.45&808.04 \\

{EBQ-G}$^\dagger$ & 945.35 & 958.63& 973.70& 757.06&\textbf{741.82} &\textbf{746.26} &\textbf{749.71}&\textbf{692.60}&\textbf{688.77}&806.00\\
 
{EBQ-N}$^\dagger$ & \textbf{934.35}& 954.64&924.31 &758.69 & 764.78& 773.74&\textbf{749.71}&692.89&695.95&\textbf{805.45} \\
 
{EBQ-GN}$^\dagger$ &1137.67& 1020.94& 1003.99& 782.51& 820.10& 750.87&832.52 &714.48&723.65&865.19\\

{EPPO} &1055.35& 1096.94& 1097.36& 767.79& 930.89& 928.59&837.90 &765.63&777.77&917.58\\

\bottomrule 
\end{tabular} 
\end{subtable}

\vspace{-2mm} \label{table:test}
\end{table*}

\subsection{Performance for high-level algorithms} \label{res:high-level}

This section summarizes various high-level algorithms across the training, evaluation (at training intervals), and testing stages. We also assess zero-shot performance and visualize the degree of improvement. For comparison, all high-level algorithms employ the same low-level SAP algorithm.
\begin{figure*}[htb] 
\centering	
	\includegraphics[width=0.95 \linewidth, height=0.65\linewidth]{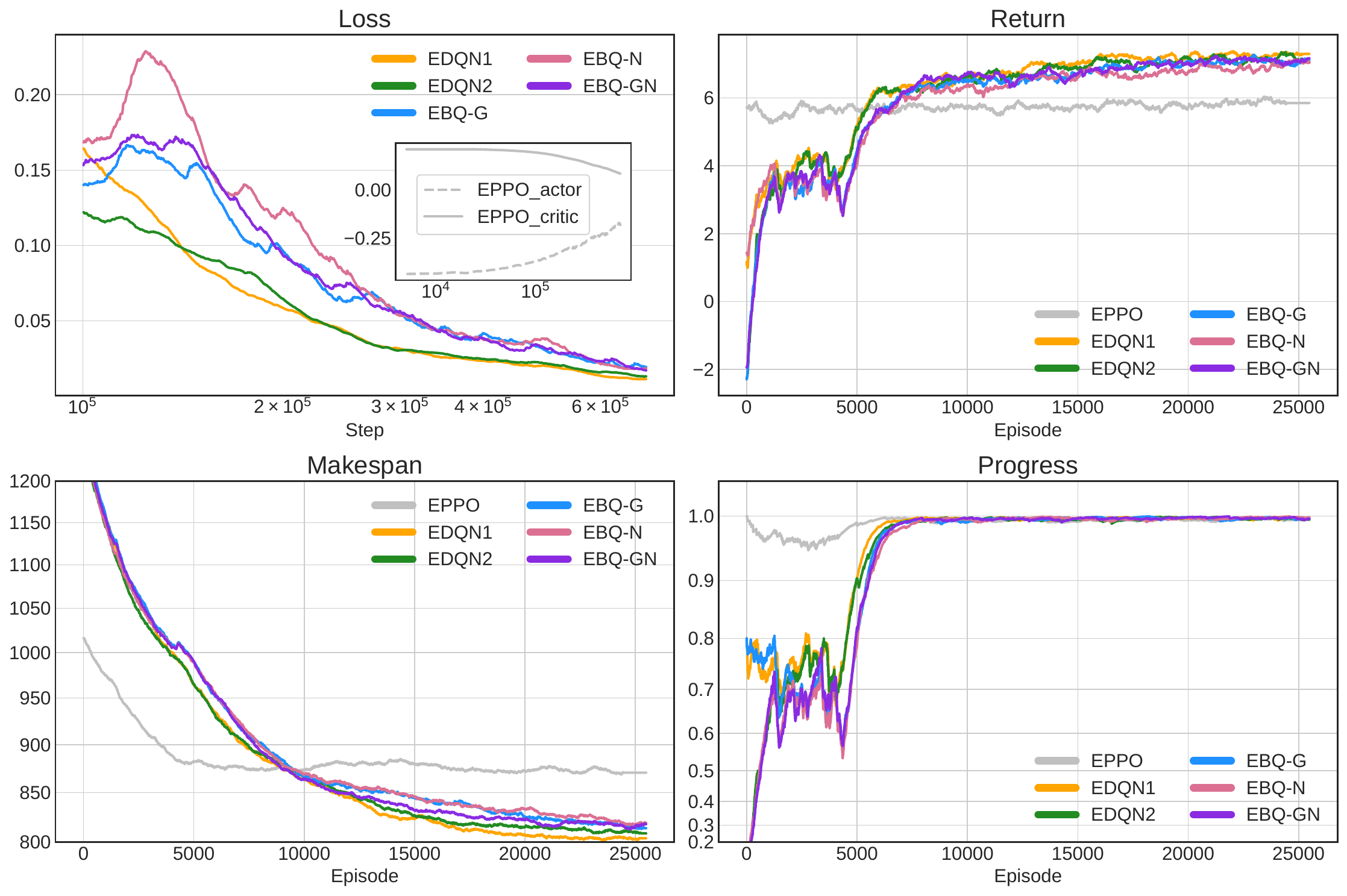}
\caption{Training curves of high-level algorithms, all use SAP as the low-level algorithm.\label{fig:metric1}} 
\vspace{-1mm}
\end{figure*}

\begin{figure*}[htb] 
\centering	
	\includegraphics[width=0.9 \linewidth, height=0.6\linewidth]{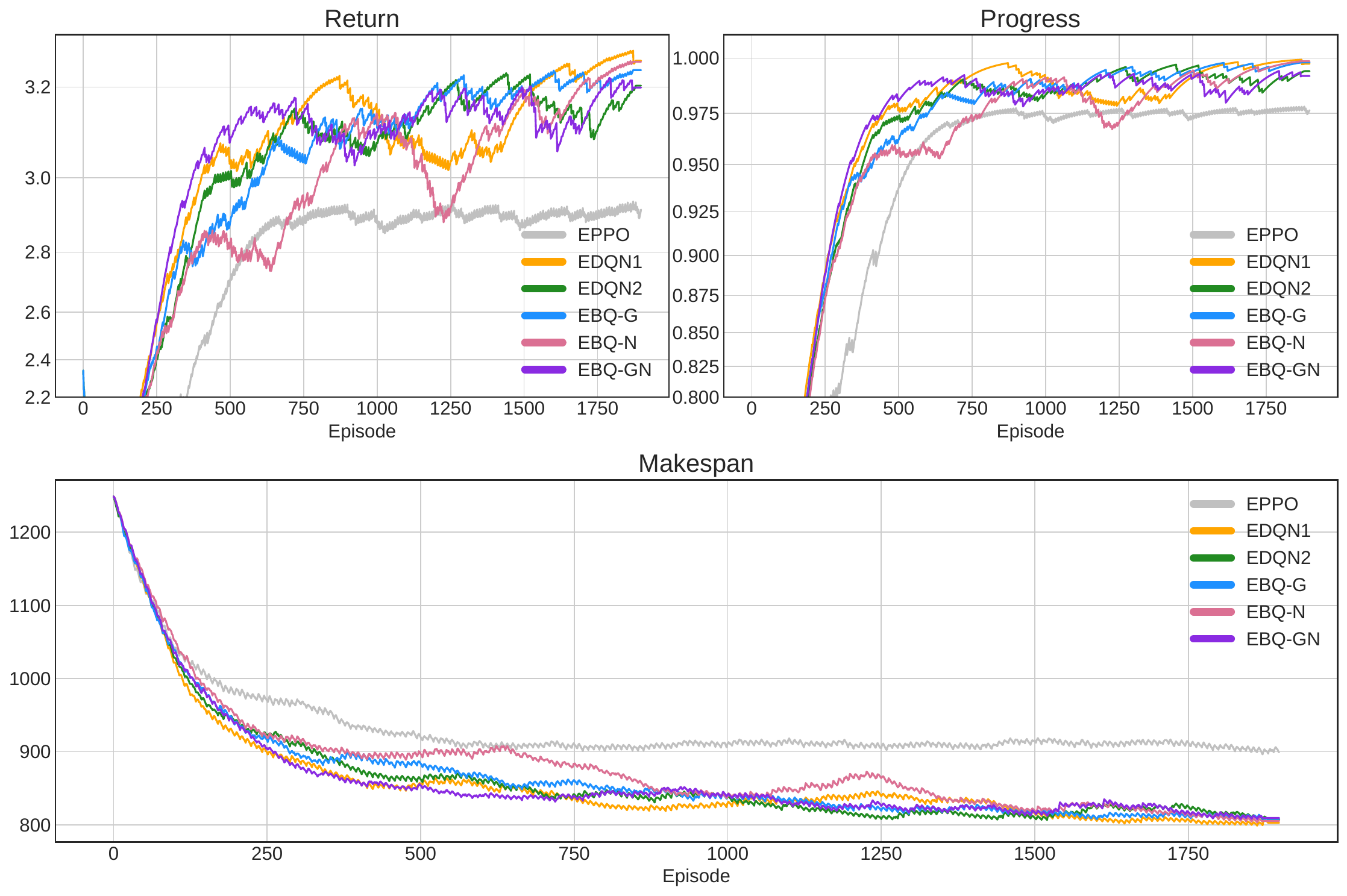}
\caption{Evaluation of high-level algorithms during training, all use SAP as the low-level algorithm.\label{fig:metric2}} 
\end{figure*}

\begin{figure*}[tpb] 
  
\centering	
	\includegraphics[width=0.99 \linewidth, height=0.53\linewidth]{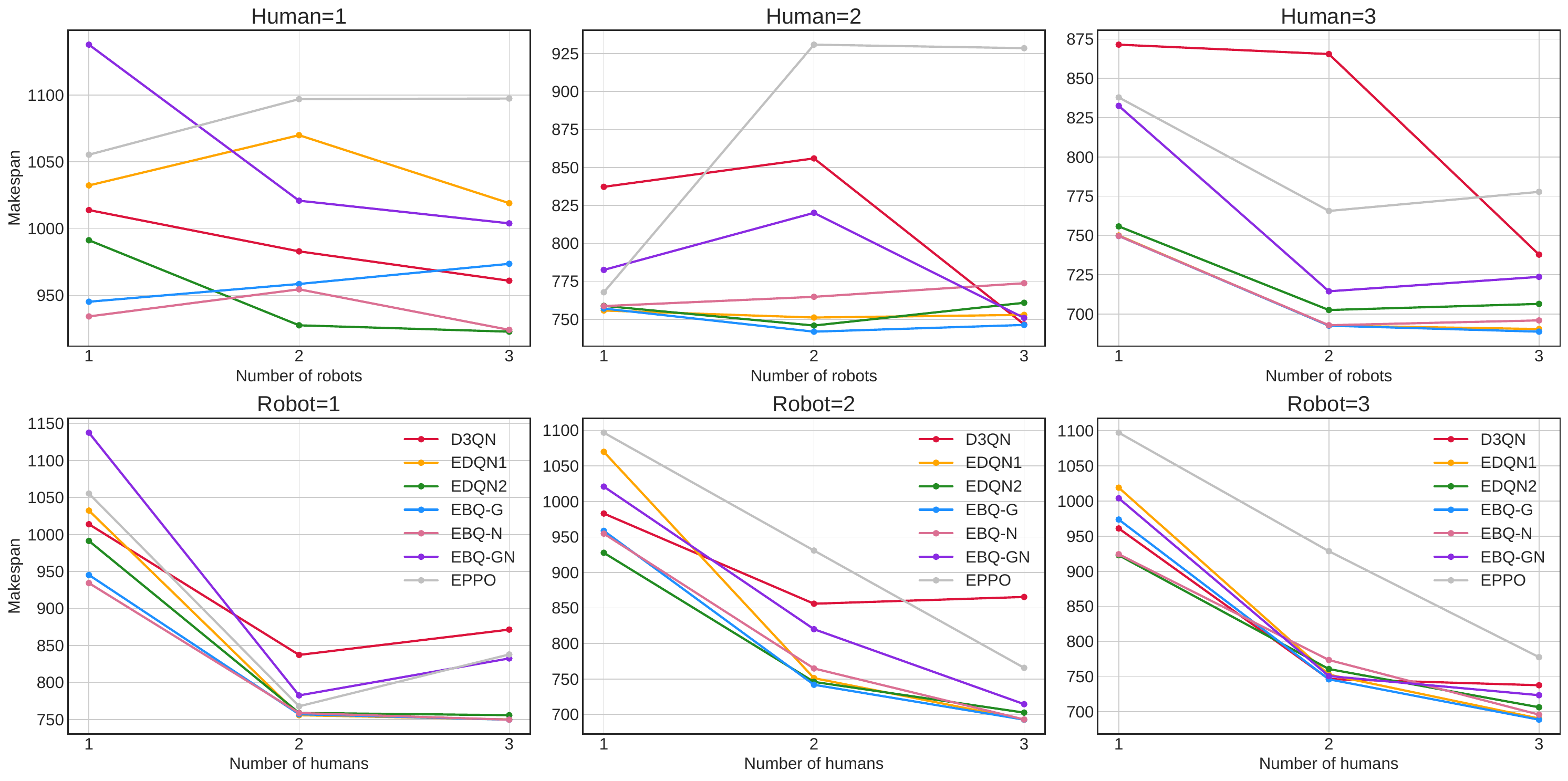}
\caption{Visualized results of Table \ref{table:test}. Makespan under varying numbers of humans and robots.\label{fig:human-robot}} 
\vspace{-2mm}
\end{figure*}
\subsubsection{Performance in training stage} \label{res:trainig}
Fig. \ref{fig:metric1} presents the training curves. The makespan and return curves indicate that EPPO has a significant gap compared to the other five algorithms, while EDQN1 achieves the best makespan upon convergence. 
Fig. \ref{fig:metric2} displays the evaluation results at training intervals. Performance generally improves over time, though EPPO is consistently the worst in all metrics, with a makespan of around 900. While there is minimal difference among the other algorithms, EBQ-N shows slight fluctuations during training, with the makespan converging around 813. 
In conclusion, EPPO demonstrates significant performance gaps across all metrics. In contrast, the other five algorithms can complete production orders and shorten makespan as training converges. 
\subsubsection{Performance in test stage} \label{res:test}
Table \ref{table:test} presents the mean makespan performance during testing. The success rate metric is omitted since all algorithms almost completed the production orders within the predefined duration. The testing was conducted with the number of humans and robots ranging from 1 to 3, with 100 trials for each configuration. 
The results indicate that EBQ-G performs best when the number of humans and robots is 2 or greater, and the mean makespan is 806. EBQ-N demonstrates superior performance in scenarios where the number of humans or robots is 1. 
When averaging the performance of all settings, EBQ-N achieves the best mean makespan, is 805.45. However, EPPO performs the worst in the mean makespan metric, is 917.58. EBQ-GN, which combines Noisy Net and epsilon-greedy exploration, exhibits suboptimal performance. In summary, D3QN and EPPO consistently perform the worst among the tested algorithms. EPPO, while favored for complex tasks due to its advanced policy optimization, requires careful tuning and strategy development \cite{de2024comparative}. In contrast, EBQ, the Q-learning-based method, is better suited for the discrete nature of the TPA problem and demonstrates superior performance.

Fig. \ref{fig:human-robot} visualizes the results of Table \ref{table:test}, displaying the impact of varying the number of humans and robots on makespan. The results show that increasing the number of humans significantly reduces the makespan. While adding more robots also leads to a general decreasing trend in makespan, the effect is less pronounced compared to adding humans. We explain the reason in Section \ref{sec:case}, as the Gantt chart highlights high idle time for robots during the production process. This is primarily due to the limited task capabilities of the robots, as they mainly assist humans rather than performing tasks independently.
The configuration of 2 humans and 2 robots is identified as the most suitable for the workspace in this study, as adding more humans or robots offers diminishing improvement.

\subsubsection{Zero-shot Performance} \label{res:zero}

\begin{figure*}[htp] 
  
\centering	
	\includegraphics[width=0.99 \linewidth, height=0.25\linewidth]{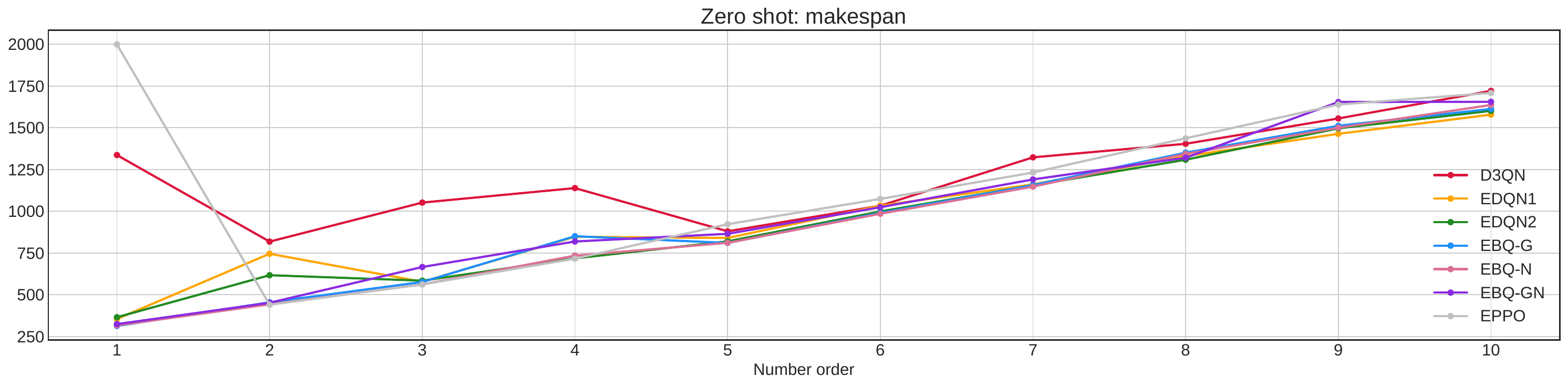}
\caption{Visualization of Table \ref{table:zero_shot}. Makespan in unseen product quantities. \label{fig:polyline2}} 
\vspace{-1mm}
\end{figure*}

\begin{table*}[htp] \label{table:zero_shot}
\vspace{-2mm}
\caption{The table presents zero-shot results, evaluating algorithms on unseen product quantities post-training. "NumN" means producing N products. The performance of our proposed model is marked with a ``$^\dagger$".}

\begin{subtable}[ht]{\linewidth}
\centering
\begin{tabular}{b{1.6cm}<{\centering}m{3em}m{3em}m{3em}m{3em}m{3em}m{3em}m{3em}m{3em}m{3em}m{3em}m{3em}}
\toprule

\multirow{2}{*}[-0.5em]{\makecell[c]{\vspace{0.1cm}Algorithm \\}} & \multicolumn{11}{c}{\normalsize \texttt{Metric: Makespan}}\\

\cline{2-12} \\[-2mm]
& \text{Num1} & \text{Num2} & \text{Num3} & \text{Num4}& \text{Num5}& \text{Num6}& \text{Num7}& \text{Num8}& \text{Num9}& \text{Num10}&\text{Mean}\\

\cline{1-12} \\[-0.5em]  
{D3QN} & 1336.32 & 817.98  & 1051.68 & 1138.52& 879.73& 1032.91& 1322.52 & 1403.76&1555.34&1720.59&1225.94\\

{EDQN1}$^\dagger$ & 354.78 & 745.46 & 577.41& 846.89& 839.91& 1033.08 & 1160.78&1329.76&1463.38& 1578.36&992.98\\

{EDQN2}$^\dagger$ & 365.21 & 616.91 & 584.10& 717.67& 819.80&998.68 &1155.13&\textbf{1308.30}&\textbf{1496.31}& \textbf{1600.91}&966.30\\


{EBQ-G}$^\dagger$ & \textbf{312.91} & 454.68 & 576.31& 850.27& \textbf{810.13}& 989.74&1156.67&1350.73&1511.72&1612.56&962.57\\
 
{EBQ-N}$^\dagger$ & 316.42 & {442.59} & {561.81} & 733.99& 810.63& \textbf{984.82} & \textbf{1147.13}&1343.44&1499.92& 1636.28&\textbf{947.70}\\
 
{EBQ-GN}$^\dagger$ & 324.00 & 451.73 & 665.94 & 818.73& 864.74& 1023.90&1190.40&1320.18&1653.77&1655.14&996.86\\


{EPPO}$^\dagger$ & 1999.00 & \textbf{440.02} & \textbf{561.62} & \textbf{715.97}& 922.31& 1073.69&1231.68&1436.62&1638.61&1708.29&1172.73\\

\bottomrule 
\end{tabular} 
\end{subtable}

\begin{subtable}[ht]{\linewidth}
\centering
\begin{tabular}{b{1.6cm}<{\centering}m{3em}m{3em}m{3em}m{3em}m{3em}m{3em}m{3em}m{3em}m{3em}m{3em}m{3em}}

\multirow{2}{*}[-0.5em]{\makecell[c]{\vspace{0.1cm}Algorithm \\}} & \multicolumn{11}{c}{\normalsize \texttt{Metric: Success Rate}}\\

\cline{2-12} \\[-2mm]
& \text{Num1} & \text{Num2} & \text{Num3} & \text{Num4}& \text{Num5}& \text{Num6}& \text{Num7}& \text{Num8}& \text{Num9}& \text{Num10}&\text{Mean}\\

\cline{1-12} \\[-0.5em]  
{D3QN} & 0.39 & 0.76 & 0.67 & 0.67& 1.00& 1.00& 0.95& 1.00 & 1.00& 1.00&0.843 \\

{EDQN1}$^\dagger$ & 0.99 & 0.81& 1.00& 0.89& 1.00&1.00 & 1.00& 1.00 & 1.00& 1.00&0.969\\

{EDQN2}$^\dagger$ & 0.98&0.89&1.00& 1.00& 1.00& 1.00&1.00& 1.00& 1.00& 1.00&0.987\\


{EBQ-G}$^\dagger$ & 1.00 & 1.00&1.00& 0.89& 1.00& 1.00&1.00& 1.00& 1.00& 1.00&0.989\\
 
{EBQ-N}$^\dagger$ & \textbf{1.00} & \textbf{1.00}& \textbf{1.00}& \textbf{1.00}& \textbf{1.00} &\textbf{1.00} &\textbf{1.00}& \textbf{1.00}& \textbf{1.00}& \textbf{1.00}&\textbf{1.000} \\
 
{EBQ-GN}$^\dagger$ & 1.00& 0.99& 0.91& 0.90& 0.99& 1.00 &1.00& 1.00& 0.99& 1.00&0.980\\


{EPPO}$^\dagger$ & 0.00 & 1.00 & 1.00 & 1.00& 1.00& 1.00&1.00&1.00&1.00&0.98&0.900\\

\bottomrule 
\end{tabular} 
\end{subtable}
 \label{table:zero_shot}
\end{table*}

During training, we set a fixed number of products to be produced, which requires producing five products for simplicity. To evaluate whether our method can perform well in unseen environments, we vary the required number of products from 1 to 10 and use makespan and success rate as metrics. Each order setting is tested 900 times, with the number of humans and robots ranging from 1 to 3. 

Table \ref{table:zero_shot} presents the zero-shot results. D3QN performs the worst, often failing to complete orders on time, resulting in low success rates (0.843) and high makespan (1225.94). EPPO also performs poorly, indicating poor adaptability to unseen scenarios.
EDQN2 achieves good results for high-product-count settings but struggles in low-product-count scenarios. The EBQ-$*$ family of algorithms consistently outperforms others, achieving better average makespan across all settings. Among them, EBQ-N demonstrated the best or near-best performance in every configuration, achieving the highest success rates (1.00) and the lowest average makespan (947.70). Specifically, EBQ-N achieved the best makespan for order quantities 2, 3, 6, and 7. For other order sizes, EBQ-N does not have the shortest makespan, but the gap was small. As a result, EBQ-N had the shortest average makespan.

Fig. \ref{fig:polyline2} visualizes the results from Table \ref{table:zero_shot}, illustrating the linear relationship between product counts and makespan. The EBQ-$*$ family of algorithms consistently achieves a 100 percent success rate as the number of products increases, demonstrating the zero-shot capabilities of our method.

\subsubsection{Performance improvement degree} \label{res:improvement}

\begin{figure*}[htp] 
\centering	
	\includegraphics[width=0.99 \linewidth, height=0.35\linewidth]{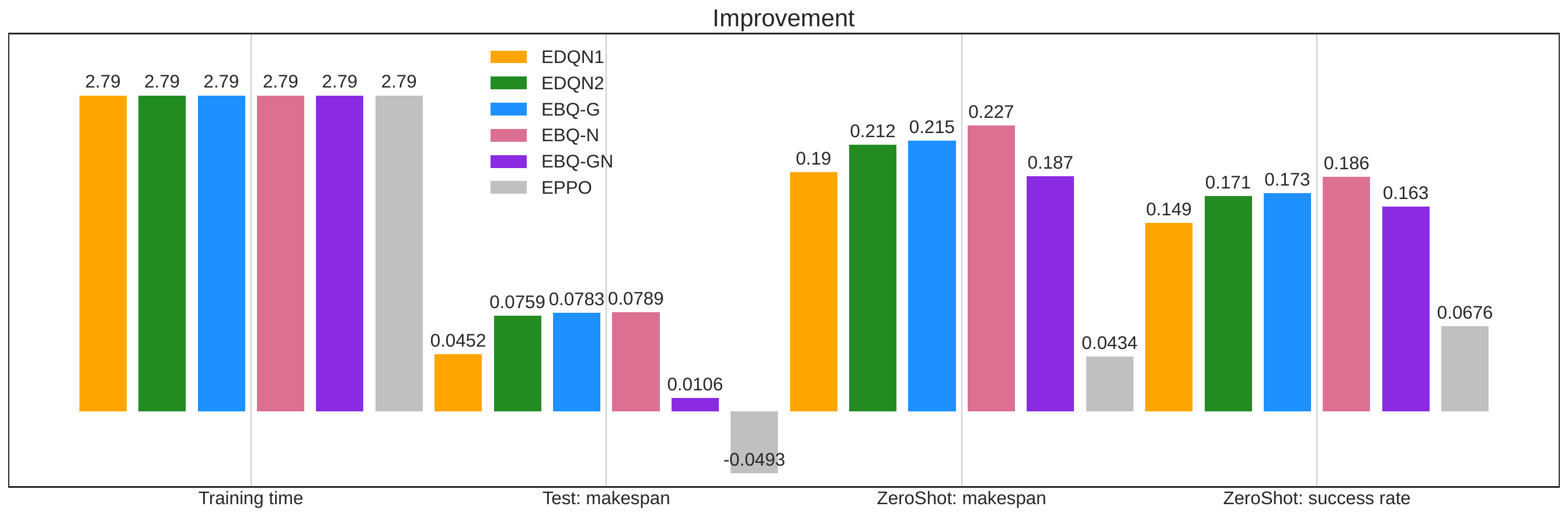}
\caption{The improvement degree of EBQ-$*$ Vs. D3QN and others, D3QN is the baseline. \label{fig:improve}}
\vspace{-1mm}
\end{figure*}

Figure \ref{fig:improve} illustrates performance improvements across four perspectives. First, the efficient training buffer mechanism significantly reduced training time, achieving a relative improvement factor of 2.79, demonstrating its effectiveness in accelerating learning. Second, test results showed notable improvements in makespan metric for all algorithms except EPPO. Third, zero-shot performance improved significantly over D3QN, with EBQ-N, leveraging Noisy Nets and a dueling network, showing the best generalization, followed by EBQ-G, which uses a $\epsilon$-greedy exploration strategy. Lastly, EBQ-GN, combining Noisy Nets and greedy exploration, underperformed, likely due to increased randomness causing training instability.

\subsection{Performance for low-level algorithms} \label{res:low-level}

\begin{figure*}[htb] 
\centering	
	\includegraphics[width=0.99 \linewidth, height=0.48\linewidth]{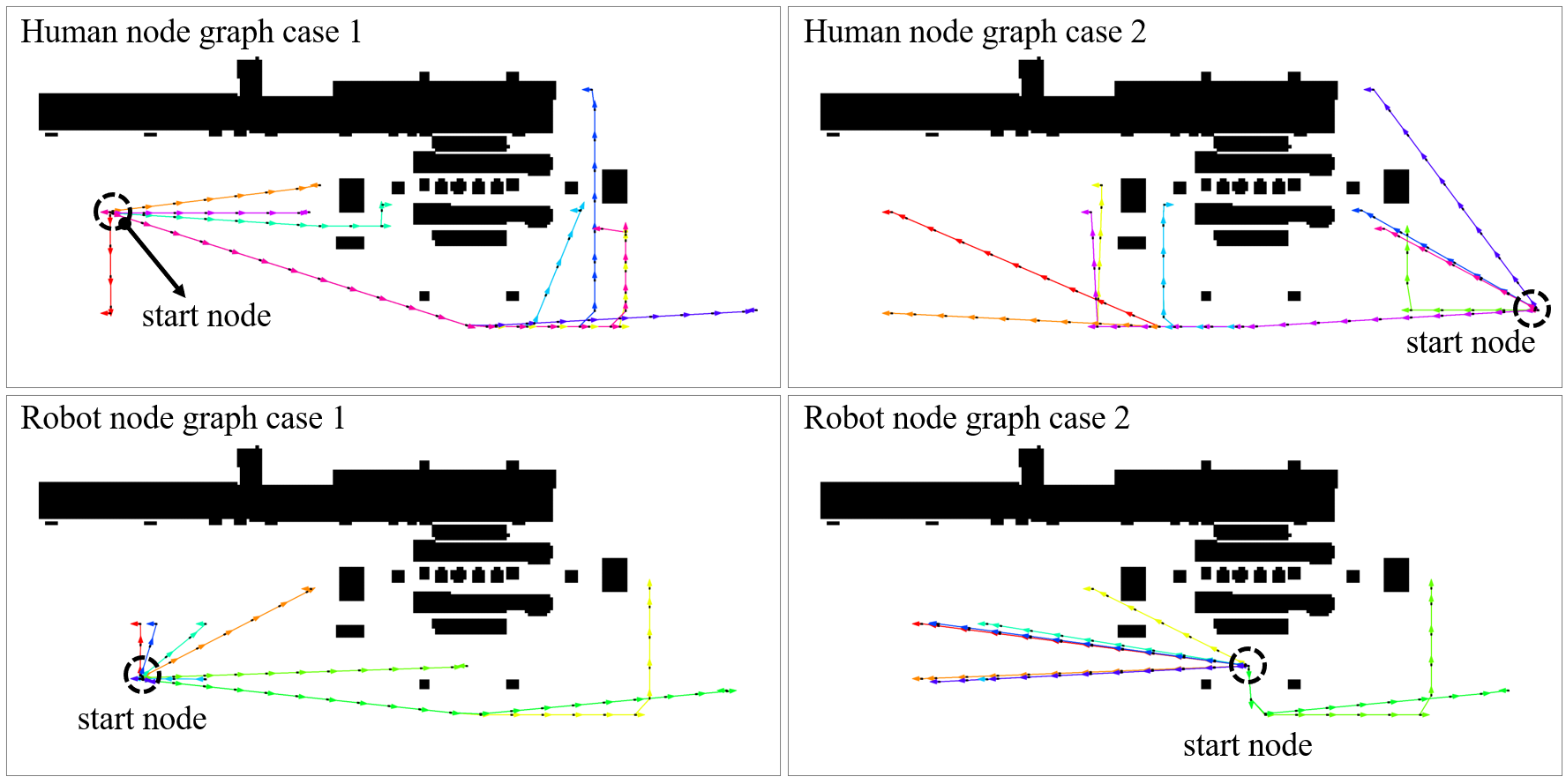}
\caption{Showcase of path planning for offline node graph generation. For both humans and robots, we present two cases each, using diverse start nodes from task-related working area nodes. Path results are provided from each start node to all other task-related working area nodes.\label{fig:node_graph_case}} 
\end{figure*}

\begin{figure*}[htb] 
\centering	
	\includegraphics[width=0.95 \linewidth, height=0.28\linewidth]{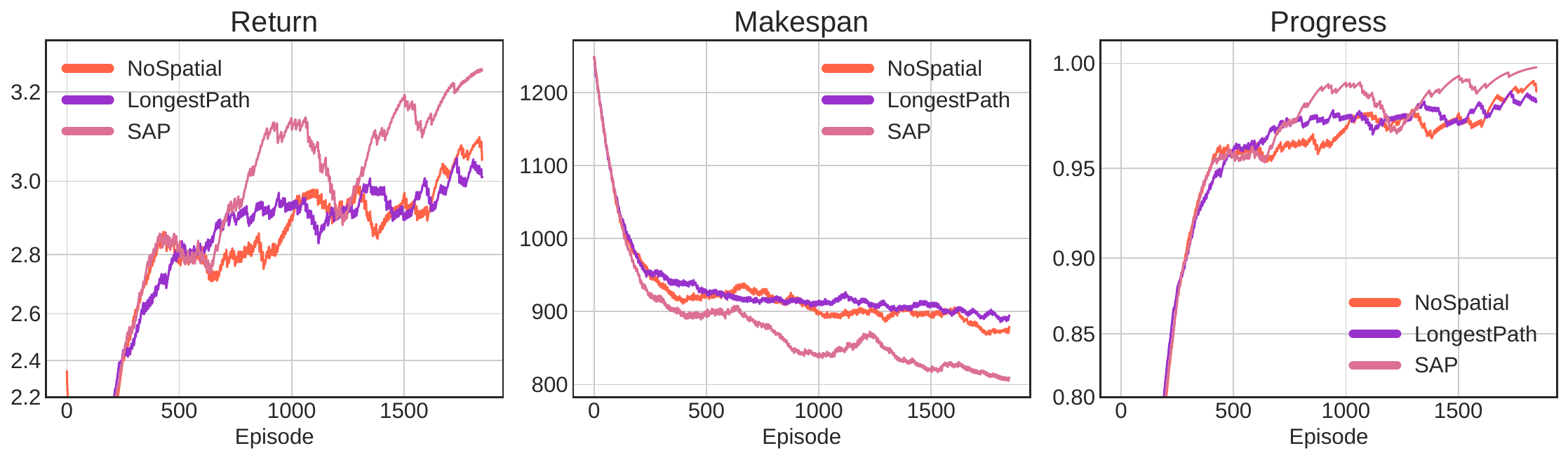}
\caption{Evaluation curves during training intervals for EBQ-N combined with various low-level algorithms.\label{fig:low_level_metric}} 
\vspace{-1mm}
\end{figure*}

\begin{table*}[ht] \label{table:test_SAP}
\caption{Makespan and distance performance in the testing phase for EBQ-N combined with various low-level algorithms. Note that a "$^\dagger$" indicates the performance of our proposed algorithm. "Hn" denotes the number of humans is set to n, while "Rn" indicates n robots during testing. }

\begin{subtable}[h]{\linewidth}
\centering
\begin{tabular}{b{1.4cm}<{\centering}m{3em}m{3em}m{3em}m{3em}m{3em}m{3em}m{3em}m{3em}m{3em}m{3em}}
\toprule

\multirow{2}{*}[-0.5em]{\makecell[c]{\vspace{0.1cm}Algorithm \\}} & \multicolumn{10}{c}{\normalsize \texttt{Metric: Distance}}\\

\cline{2-11} \\[-2mm]
& \text{H1,R1} & \text{H1,R2} & \text{H1,R3} & \text{H2,R1}& \text{H2,R2}& \text{H2,R3}& \text{H3, R1}& \text{H3,R2}& \text{H3,R3}& \text{Mean}\\

\cline{1-11} \\[-0.5em]  
{LongestPath} & 586.82 & 684.18& 748.2& 612.86& 719.35& 767.67&627.43&763.51&801.44 & 874.73 \\

{NoSpatial} &\textbf{623.82}& 686.40&670.53&572.23&675.65&714.49&574.96&666.99&706.43&654.61 \\
 
{SAP}$^\dagger$ & 643.82& \textbf{613.19}& \textbf{605.89} &\textbf{536.63} & \textbf{561.16}& \textbf{582.44}&\textbf{502.36}&\textbf{496.71}&\textbf{499.27}&\textbf{560.16} \\

\bottomrule 
\end{tabular} 
\end{subtable}
\begin{subtable}[h]{\linewidth}
\centering
\begin{tabular}{b{1.4cm}<{\centering}m{3em}m{3em}m{3em}m{3em}m{3em}m{3em}m{3em}m{3em}m{3em}m{3em}}
\toprule

\multirow{2}{*}[-0.5em]{\makecell[c]{\vspace{0.1cm}Algorithm \\}} & \multicolumn{10}{c}{\normalsize \texttt{Metric: Makespan}}\\

\cline{2-11} \\[-2mm]
& \text{H1,R1} & \text{H1,R2} & \text{H1,R3} & \text{H2,R1}& \text{H2,R2}& \text{H2,R3}& \text{H3, R1}& \text{H3,R2}& \text{H3,R3}& \text{Mean}\\

\cline{1-11} \\[-0.5em]  
{LongestPath} &973.35& 1064.33 & 1051.72  & 825.04& 767.75& 861.32& 764.46&840.96&804.07 & 883.67 \\

{NoSpatial} &983.35& 1181.39&995.71&794.91&798.23&839.59&803.93&812.29&785.00&888.23 \\
 
{SAP}$^\dagger$ & \textbf{945.35} & \textbf{958.63}& \textbf{973.70}& \textbf{757.06}&\textbf{741.82} &\textbf{746.26} &\textbf{749.71}&\textbf{692.60}&\textbf{688.77}&\textbf{806.00} \\

\bottomrule 
\end{tabular} 
\end{subtable}

\vspace{-2mm} \label{table:test_SAP}
\end{table*}

Section \ref{sec:SAP} details the SAP algorithm, which comprises two stages: offline node graph generation and real-time task allocation. The visualization results of offline node graph generation are presented first, while real-time task allocation is showcased in Section \ref{sec:case}. Figure \ref{fig:node_graph_case} illustrates path planning for offline node graph generation, featuring two cases each for humans and robots, using diverse start nodes from task-related working area nodes. Path results from each start node to all other task-related working area nodes are recorded to construct the final node graph. During real-time task allocation, we update the closest node for each human or robot based on their real-time global position and calculate distances to various task-related working areas by directly extracting precomputed path planning results from the node graph.

For real-time task allocation, Fig. \ref{fig:low_level_metric} presents evaluation curves during training intervals for EBQ-N combined with various low-level algorithms. SAP, leveraging the precomputed offline node graph, outperforms NoSpatial and LongestPath, particularly in the makespan metric. Table \ref{table:test_SAP} summarizes test-stage performance across algorithms. LongestPath yields the longest movement distance by consistently selecting the farthest entities. NoSpatial, lacking spatial awareness and relying on random allocation, produces the second-longest distance. In contrast, our SAP algorithm, which allocates tasks to humans and/or robots based on the closest distance paradigm, achieves the shortest distance metric. Consequently, SAP exhibits a significant performance advantage in makespan over the other two algorithms. When combined with a high-level algorithm, SAP enables effective real-time task allocation in dynamic production environments, where human and robot movement within the workspace during production is a critical factor.

\subsection{Case study} \label{sec:case}

\begin{figure*}[thb] 
\centering	
	\includegraphics[width=0.99 \linewidth, height=1.3\linewidth]{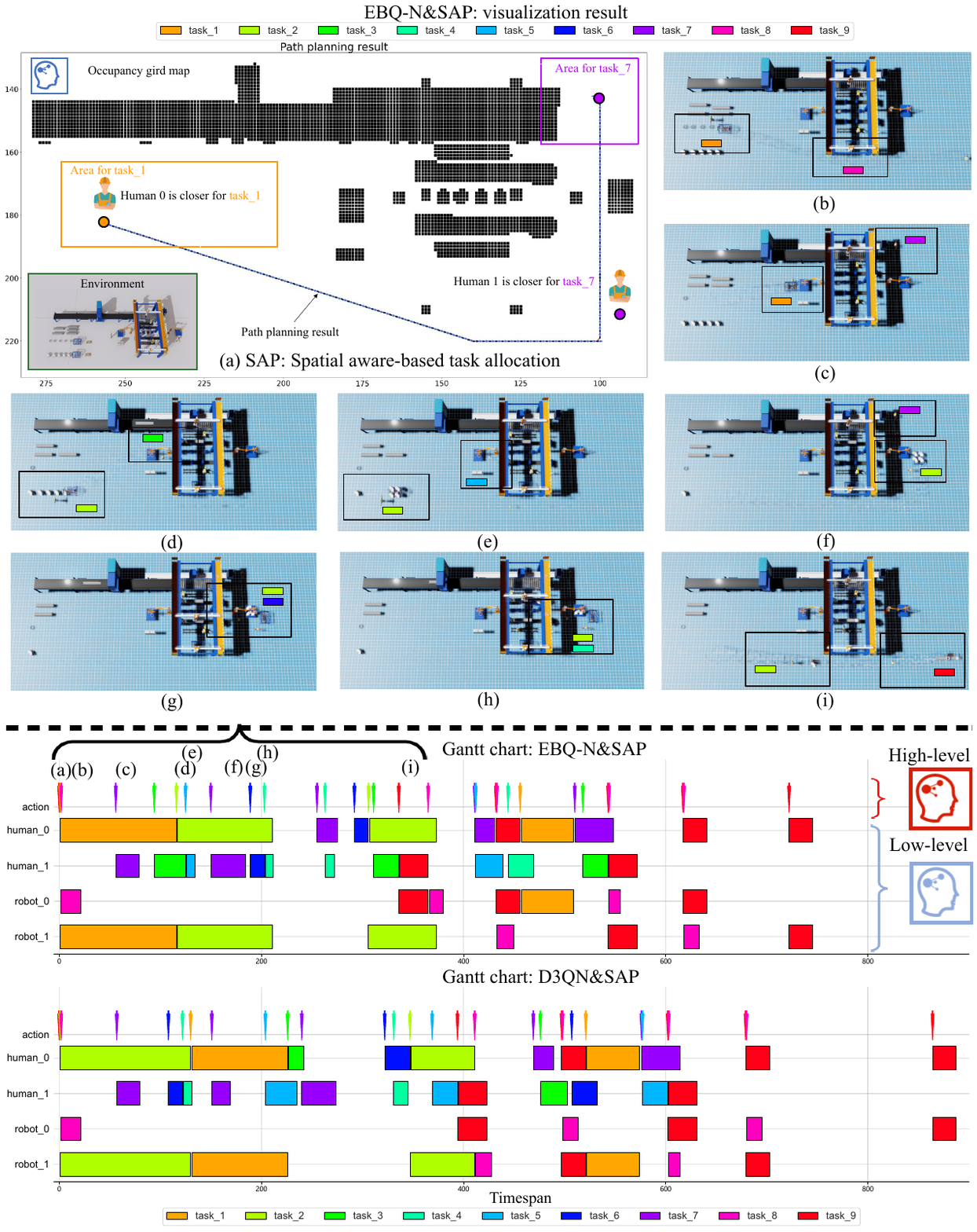}
\caption{Case study: Gantt chart and visualization results. Tasks' details in Fig. 
\ref{fig:dependency_and_task}, and environment's details in Fig. \ref{fig:environment}.\label{fig:case}} 
\end{figure*}

Fig. \ref{fig:case} illustrates the real-time human-robot TPA results using EBQ$\&$SAP, with task details provided in Fig. \ref{fig:dependency_and_task} and environmental details shown in Fig. \ref{fig:environment}. The upper half presents snapshots of the production process using the EBQ-N$\&$SAP algorithm, which achieves the best average performance across all metrics. The lower half displays the corresponding Gantt chart results, including a baseline comparison with D3QN. 
To detail the process, subfigure (a) visualizes the initial path planning-based task allocation method, followed by environmental snapshots from subfigures (b) to (i). We use rectangular frames and color-coded task bars in each subfigure to represent different tasks' working areas, making the visualizations easier to understand. Each snapshot is marked on the Gantt chart, with corresponding time frames labeled from (a) to (i) for clarity.
\subsubsection{Real-time human-robot TPA process by EBQ$\&$SAP}
We begin by describing each subfigure from (a) to (i) concerning EBQ-N's Gantt chart.
(a) At the beginning of the production process, the high-level agent gathers state information and selects task 1, which involves conveying a flange to the side storage area. The lower-level agent, SAP, allocates the tasks and subtasks to humans and robots. SAP generates an occupancy map for path planning, determining that human 0 and robot 1 are suitable for executing task 1. Subtask sequences of task 1 are subsequently generated for both human 0 and robot 1.
(b) The high-level agent selects task 8 as the next action. This task involves a robot conveying the cage to the product collection area and does not require human involvement. Meanwhile, human 0 and robot 1 continue executing the subtasks of task 1.
(c) Human 1 is allocated task 7, which involves selecting and activating the controlling code for workstations.
(d) Human 0 begins task 2, which involves conveying bending ducts to the side storage area. Simultaneously, human 1 starts task 3, loading a flange onto laser welding workstation 1.
(e) Human 0 continues task 2, while human 2 begins task 5, loading a flange onto welding workstation 2.
(f) Human 0 continues task 2, while human 1 resumes task 7, selecting and activating the product processing code for workstations.
(g) Human 0 persists with task 2, and human 1 starts task 6, which involves loading a bending duct onto welding workstation 2.
(h) Human 0 continues task 2, and human 1 initiates task 4, loading a bending duct onto welding workstation 1.
(i) After some time, the first product is produced and then collected by the cage. Human 1 and robot 0 are then allocated task 9, which involves conveying the product to the storage area.
Finally, a similar TPA process is conducted till finished the production order. 
\subsubsection{EBQ-N Vs. D3QN}
The Gantt chart results indicate that D3QN performs worse than EBQ-N, with a gap of approximately 100 time steps. Although real-time human-robot TPA is complex and highly dynamic, some notable differences can be observed. For instance, at the beginning of the process, EBQ-N prioritizes completing task 1 before task 2, whereas D3QN chooses to finish task 2 first, which extends the overall production time. Additionally, EBQ-N demonstrates higher utilization of human and robot resources compared to D3QN, resulting in shorter idle intervals. We also observed that human resources are more critical than robots, as humans possess more diverse capabilities. Robots are for assistance with human work; consequently, robots experience longer idle periods.

In conclusion, the on-time accomplishment of the production order demonstrates the effectiveness of our EBQ$\&$SAP algorithm, which decomposes intricate and dynamic problems into a two-level decision-making process. For the high-level agent, the superior performance of EBQ-N compared to D3QN highlights the advantages of our design methodology. At the low level, the SAP method effectively incorporates spatial information through path planning while efficiently handling task allocation.

\section{Discussion}
The proposed efficient buffer-based deep Q-learning with a path planning-based spatially aware (EBQ$\&$SAP) algorithm is a novel approach for the human-robot TPA problem. Compared to prior work, EBQ$\&$SAP demonstrates several advancements: (1) for the RL-based high-level agent, EBQ reduces training time fourfold and achieves notable test performance improvements over D3QN \cite{wang2016dueling}, which only utilizes prioritized experience replay \cite{schaul2015prioritized, pan2022understanding}, by incorporating an efficient buffer-based training strategy, and surpass EPPO \cite{schulman2017proximal}, which requires more careful tuning and strategy development; (2) the EBQ-N network, integrating dueling networks \cite{wang2016dueling} and Noisy Nets \cite{NoisyNet}, outperforms EDQN2 \cite{van2016deep} in zero-shot makespan and success rate, highlighting the effectiveness of our architecture design; and (3) the path-planning-based low-level agent, SAP, unlike NoSpatial, which lack spatial awareness, excels in makespan by leveraging real-time spatial information to handle dynamic production scenarios.

Despite these advancements, our study has limitations. First, real-world implementation is hindered by constraints in facilities and digital infrastructure. Deploying the algorithm requires real-time integration of multi-source production data from humans, robots, machines, and environments, as well as feedback on TPA decisions to guide production. This necessitates a robust digital twin infrastructure, which is challenging to implement currently. Second, task and subtask decomposition, typically customized for specific production processes, is difficult to standardize and requires predefined structures \cite{cheng2019task}. This manual process is time-consuming and lacks flexibility across diverse production lines with varying task natures, machines, and environments.

\section{Conclusion}

In this work, we address the human-robot TPA problem in a complex and dynamic manufacturing environment characterized by multiple humans and robots, as well as real-time spatial information. To address these challenges, we propose EBQ$\&$SAP, a hierarchical human-robot algorithm designed to enhance collaboration and effectively reduce the complexity of the problem. 
The algorithm consists of two main components: EBQ and SAP. EBQ acts as a high-level decision agent that selects high-level tasks in real-time. SAP acts as a low-level agent, performing path planning that considers spatial factors and allocating tasks to both humans and robots. 
For the EBQ algorithm, we enhance the standard D3QN by introducing an efficient training buffer mechanism to address the sparse reward problem inherent in the long-running nature of production processes. This enhancement significantly increases training efficiency and overall performance.
Regarding network design, we employ a Transformer architecture to handle large amounts of heterogeneous information effectively. In addition, we integrate a dueling network to enhance the model's capability and use Noisy Net to improve the exploration strategy. The experimental results demonstrate the effectiveness of our proposed algorithm and its novelties. 

However, several topics remain open for future research. First, the application of our algorithm to a wider range of manufacturing scenarios should be explored to enhance its adaptability and robustness. Second, real-world testing faces challenges due to limited facilities and digital technology infrastructure. Combining our algorithm with digital twin technology is a promising direction to address these limitations and facilitate real-world implementation. Third, incorporating large language or vision models offers significant potential due to their versatility and advanced capabilities in handling complex information. 
Additionally, automating task and subtask generation and decomposition is a critical direction to improve efficiency during the task description stage.
Finally, integrating safety considerations is essential, as ensuring human and enterprise safety is a fundamental requirement in production environments.
\section*{CRediT authorship contribution statement}
\textbf{Jintao Xue}: Methodology, Software, Data curation, Writing – original draft. \textbf{Xiao Li}: Conceptualization, Supervision, Writing – review $\&$ editing. \textbf{Nianmin Zhang}: methodology analysis, experiment design.

\section*{Declaration of Competing Interest} The authors declare that they have no known competing financial interests or personal relationships that could have appeared to influence the work reported in this paper. 

\section*{Data availability} Data will be made available on request.

\section*{Acknowledgments} 
The work described in this paper is supported by grants from Technology Cooperation Funding Scheme (TCFS) (Ref No.GHP/321/22SZ), The University of Hong Kong (Ref No.109002002), and Innovation and Technology Fund (ITF) (Ref No. TP/041/24LP).

\bibliographystyle{cas-model2-names}

\bibliography{cas-refs}

\end{document}